\documentclass{article}

\usepackage{amsmath,amsfonts,bm}

\def\eqref#1{equation~\ref{#1}}

\def\1{\bm{1}}

\DeclareMathAlphabet{\mathsfit}{\encodingdefault}{\sfdefault}{m}{sl}
\SetMathAlphabet{\mathsfit}{bold}{\encodingdefault}{\sfdefault}{bx}{n}

\newcommand{\E}{\mathbb{E}}

\usepackage[colorlinks=true]{hyperref}
\hypersetup{%
,urlcolor=blue
,citecolor=blue
,linkcolor=red
}

\usepackage{microtype}
\usepackage{graphicx}
\usepackage{subfigure}
\usepackage{booktabs} %

\usepackage{hyperref}

\usepackage{wasysym}
\usepackage{url}
\usepackage{graphicx}
\usepackage{wrapfig}
\usepackage{mwe}
\usepackage{multirow}
\usepackage{adjustbox}
\usepackage{booktabs}

\newcommand{\phii}{{\phi_i}}
\newcommand{\phiiz}{{\phi_{i,0}}}
\newcommand{\psij}{{\psi_j}}
\newcommand{\psijz}{{\psi_{j,0}}}

\usepackage[accepted]{icml2024}

\usepackage{amsmath}
\usepackage{amssymb}
\usepackage{mathtools}
\usepackage{amsthm}

\usepackage[capitalize,noabbrev]{cleveref}

\theoremstyle{plain}
\newtheorem{theorem}{Theorem}[section]

\theoremstyle{definition}

\theoremstyle{remark}

\usepackage[textsize=tiny]{todonotes}

\icmltitlerunning{Multi-attribute Selective Suppression for Utility-preserving Data Transformation from an Information-theoretic Perspective}

\begin{document}

\twocolumn[
\icmltitle{MaSS: Multi-attribute Selective Suppression for Utility-preserving Data Transformation from an Information-theoretic Perspective}

\icmlsetsymbol{equal}{*}

\begin{icmlauthorlist}
\icmlauthor{Yizhuo Chen}{equal,yyy,comp}
\icmlauthor{Chun-Fu (Richard) Chen}{equal,comp}
\icmlauthor{Hsiang Hsu}{comp}
\icmlauthor{Shaohan Hu}{comp}
\icmlauthor{Marco Pistoia}{comp}
\icmlauthor{Tarek Abdelzaher}{yyy}
\end{icmlauthorlist}

\icmlaffiliation{yyy}{Department of Computer Science, University of Illinois Urbana-Champaign, USA}
\icmlaffiliation{comp}{Global Technology Applied Research, JPMorgan Chase, USA}

\icmlcorrespondingauthor{Yizhuo Chen}{yizhuoc@illinois.edu}
\icmlcorrespondingauthor{Chun-Fu (Richard) Chen}{richard.cf.chen@jpmchase.com}

\icmlkeywords{Machine Learning, ICML}

\vskip 0.3in
]

\printAffiliationsAndNotice{\icmlEqualContribution} %

\begin{abstract}
The growing richness of large-scale datasets has been crucial in driving 
the rapid advancement and wide adoption of machine learning technologies. 
The massive collection and usage of data, however,
pose an increasing risk for people's private and sensitive information
due to either inadvertent mishandling or malicious exploitation.
Besides legislative solutions,
many technical approaches have been proposed towards data privacy protection.
However, they bear various limitations such as leading to degraded data availability and utility, 
or relying on heuristics and lacking solid theoretical bases.
To overcome these limitations,
we propose a formal information-theoretic definition for this utility-preserving privacy protection problem,
and design a data-driven learnable data transformation framework
that is capable of selectively suppressing sensitive attributes from target datasets 
while preserving the other useful attributes,
regardless of whether or not they are known in advance or explicitly annotated for preservation.
We provide rigorous theoretical analyses on the operational bounds for our framework,
and carry out comprehensive experimental evaluations using datasets of a variety of modalities,
including facial images, voice audio clips, and human activity motion sensor signals.
Results demonstrate the effectiveness and generalizability of our method under various configurations on a multitude of tasks. Our code is available at \url{https://github.com/jpmorganchase/MaSS}.

\end{abstract}

\section{Introduction}
\label{sec:intro}

The recent rapid advances and wide adoption
of machine learning technologies 
are largely attributed to not only the explosive growth in raw computing power,
but also the unprecedented availability of large-scale datasets, 
for example, the monumental computer vision dataset ImageNet~\citep{deng2009imagenet}, 
the large multi-lingual web corpus~\citet{commoncrawl}, 
and the widely used UCI HAR dataset~\citep{anguita2013public}.
While the vast amount of data serves as the rich basis for machine learning algorithms to learn from,
the ubiquitous data collection and usage have drawn serious privacy concerns
since people's private and sensitive information could be leaked 
through inadvertent mishandling as well as deliberate malicious exploitation. 
Therefore, various regulatory policies, 
such as~\citet{gdpr} and~\citet{ccpa},
have been drafted and put in place to guardrail the handling and usage of data.
While such legislative solutions do generally help mitigate the privacy concerns,
they also tend to pose blanket restrictions
that result in degraded data availability.
Therefore, there has been a growing interest in developing more sophisticated, flexible technical solutions.

Towards this goal, many techniques have been proposed.
One of the most well-known studies is the protection against membership inference attacks, 
also known as Differential Privacy~\citep{dwork2014algorithmic, mironov2017renyi, abadi2016deep}. 
It focuses on preventing attackers from differentiating between two neighboring sets of samples by observing the change in the distribution of output statistics.
Another widely discussed notion of privacy is 
the protection against attribute inference attacks, 
often referred to as Information-theoretic (IT) Privacy~\citep{hukkelaas2019deepprivacy, bertran2019adversarially, huang2018generative, hsu2020obfuscation}. 
This line of work aims at transforming a dataset
to remove or suppress its sensitive attributes
while preserving its utility for downstream tasks. 
In this paper, we focus our discussion on providing IT Privacy protection.

Developing a data transformation framework for IT Privacy presents multiple challenges. 
Specifically, we identified 5 desired properties for an IT Privacy data transformation framework, which can be summarized as \textit{SUIFT}: 
1) \textit{S}ensitivity suppression: the capability to suppress annotated sensitive attributes from the dataset; 
2) \textit{U}tility preservation: the capability to preserve specifically annotated useful attributes in the dataset to facilitate downstream usage; 
3) \textit{I}nvariance of sample space: keeping the transformed data in the original space as the input data, to enable plug-in usability for pretrained off-the-shelf models and to deliver better re-usability for the community; 
4) \textit{F}eature management without annotation: the capability to manage unannotated generic features in the dataset, by either suppressing them or preserving them when they are considered either useful or sensitive.
5) \textit{T}heoretical basis: all the proposed components of the data transformation frameworks being entirely driven by a unified information-theoretic basis to ensure safety.

Various techniques have been proposed towards the goal of SUIFT in IT Privacy, as summarized in Table \ref{tb:suift}.
However, each of them is limited in missing some of the desired properties of SUIFT.
For example, \citet{bertran2019adversarially}, \citet{wu2020privacy}, and \citet{kumawat2022privacy}
can only ensure the predictability in the transformed data for attributes that have already been explicitly annotated for preservation; no considerations are given to managing data's unannotated attributes.
On the other hand, \citet{huang2018generative}, \citet{malekzadeh2019mobile} and \citet{madras2018learning} do account for unannotated attributes,
but their designs for unannotated attributes preservation are mostly heuristic-driven and lack rigorous theoretical bases, 
which could limit their applicability, especially for scenarios involving highly sensitive information.

\begin{table}[!t]\scriptsize
\caption{\small A summary of related works on IT Privacy. \CIRCLE, \Circle\xspace and \LEFTcircle\xspace indicate \textit{satisfied}, \textit{not satisfied} and \textit{partially satisfied}, respectively. S, U, I, F, and T are abbreviations of 
    \textit{Sensitivity suppression}, 
    \textit{Utility preservation}, 
    \textit{Invariance of sample space}, 
    \textit{Feature management without annotation}, and 
    \textit{Theoretical basis} respectively. 
Apart from our method, MaSS (to be introduced shortly, and described in detail in Section~\ref{sec:method}), 
none of the listed methods fully satisfy all components of SUIFT (discussed in detail in Section~\ref{sec:rw} and Appendix~\ref{app:rw}).}
\label{tb:suift}
\begin{center}
\begin{adjustbox}{max width=\linewidth}
\begin{tabular}{lccccc}
\toprule 
Method         & S & U & I & F & T \\
\midrule
DeepPrivacy \cite{hukkelaas2019deepprivacy} & \CIRCLE & \LEFTcircle & \CIRCLE & \LEFTcircle & \Circle \\
CiaGAN \cite{maximov2020ciagan} & \CIRCLE & \LEFTcircle & \CIRCLE & \LEFTcircle & \Circle \\
\citet{hsu2020obfuscation} & \CIRCLE & \Circle & \CIRCLE & \LEFTcircle & \CIRCLE \\
ALR \cite{bertran2019adversarially} & \CIRCLE & \CIRCLE & \CIRCLE & \Circle & \CIRCLE \\
PPDAR \cite{wu2020privacy} & \CIRCLE & \CIRCLE & \CIRCLE & \Circle & \Circle \\
BDQ \cite{kumawat2022privacy} & \CIRCLE & \CIRCLE & \CIRCLE & \Circle & \Circle \\
ALFR \cite{edwards2015censoring} & \LEFTcircle & \CIRCLE & \Circle & \CIRCLE & \LEFTcircle \\
LAFTR \cite{madras2018learning} & \LEFTcircle & \CIRCLE & \Circle & \CIRCLE & \LEFTcircle \\
GAP \cite{huang2018generative} & \CIRCLE & \Circle & \CIRCLE & \CIRCLE & \LEFTcircle \\
MSDA \cite{malekzadeh2019mobile} & \CIRCLE & \CIRCLE & \CIRCLE & \CIRCLE & \LEFTcircle \\
SPAct \cite{dave2022spact} & \Circle & \CIRCLE & \CIRCLE & \CIRCLE & \Circle \\

\midrule
MaSS (our method) & \CIRCLE & \CIRCLE & \CIRCLE & \CIRCLE & \CIRCLE \\
\bottomrule
\end{tabular}
\end{adjustbox}
\end{center}
\vspace{-0.1in}
\end{table}

\begin{figure}[thbp]
    \centering
    \includegraphics[width=\linewidth]{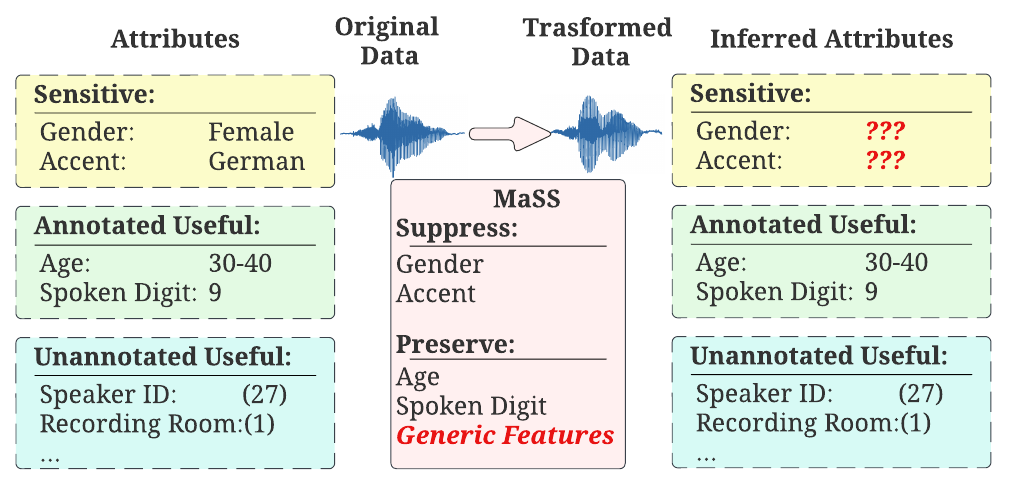}
    \caption{\small An illustrative use case of MaSS: The original data sample is a voice clip of a person speaking a digit, where its attributes ``gender" and ``accent" are considered as sensitive, while its ``age" and ``spoken digit" are annotated as useful. We are also interested in preserving generic features of the data. For example, the voice clip may contain attributes such as ``speaker ID" or ``recording room" that could prove to be useful down the road, but are not necessarily explicitly annotated yet at the time of processing. After the transformation of MaSS, sensitive attributes can no longer be accurately inferred, but the other useful attributes are preserved in the transformed data.}
    \label{fig:motivation}
\end{figure}

To address these limitations, in this paper we present MaSS,
a \textbf{M}ulti-\textbf{a}ttribute \textbf{S}elective \textbf{S}uppression framework that aims at satisfying all 5 components of SUIFT. 
Specifically, we formulate the IT Privacy as an optimization problem from the perspective of information theory, and then convert the optimization problem into a fully differentiable trainable framework parameterized by neural networks, with sound analyses on the design derivation and operational bounds. MaSS is capable of suppressing multiple selected sensitive attributes, and preserving multiple useful attributes regardless of whether they are annotated or not.
An illustrative use case of MaSS is shown in Figure \ref{fig:motivation}.
We also compare MaSS with various baselines extensively on three datasets of different modalities, 
namely voice recordings, human activity motion sensor signals, 
and facial images,
and show its practical effectiveness under various configurations.%

The contributions of this paper are summarized as follows: 1) We propose MaSS, an information theory driven data transformation framework satisfying all 5 identified desirable properties in IT Privacy, namely SUIFT; 2) We provide rigorous theoretical analyses on the design derivation and operational bounds of our proposed multi-attribute data transformation framework; and 3) We experimentally evaluate MaSS extensively on voice audio, human activity motion sensor signal, and facial image datasets, and demonstrate its effectiveness and generalizability.

Omitted proofs, details on experiment setups and training, and additional results are included in the Appendix. 

\section{Related Works}\label{sec:rw}

\paragraph{Privacy-preserving mechanisms.} 
A privacy-preserving mechanism ensures privacy by randomizing a function of data in order to thwart unwanted inferences.
There are two selections of the functions that lead to different privacy notions. 
If the function is the output of a query over a database, the privacy notion is termed differential privacy (DP) \citep{dwork2006calibrating}, which requires the results of a
query be approximately the same for small perturbations of data, and can usually be achieved by additive noise mechanisms  (e.g., Gaussian,
Laplacian or exponential noise \citep{dwork2014algorithmic,sun2020human,zhang2018differentially,abadi2016deep}).
Different from DP, if the function is a conditional distribution that anonymizes sensitive information in the data while preserving non-sensitive information, it leads to the other privacy notion called information-theoretic (IT) privacy. 
The motivation behind IT privacy is to improve the data quality after anonymization with the additional information of the utility attributes. 
See \citet{hsu2021survey} for a more detailed discussion on the two privacy notions.
Since our goal is to not only suppress the sensitive attributes but also preserve the data utility concurrently, the MaSS framework falls within the field of IT privacy.

\paragraph{IT Privacy protection for annotated attributes.} 
By reviewing related studies for IT Privacy and examining the requirements of downstream applications, we identified 5 desirable properties of IT Privacy mechanisms, namely \textit{SUIFT} as summarized in Section \ref{sec:intro} and Table \ref{tb:suift}. Nevertheless, previous studies proposed for IT Privacy are limited in certain properties. For instance, DeepPrivacy~\citep{hukkelaas2019deepprivacy} employs a CGAN, conditioned on image background and pose features, to synthesize anonymized facial images. To further ensure de-identification, CiaGAN~\citep{maximov2020ciagan} proposes to condition the CGAN on an identity control vector, creating images with fabricated identities. Nevertheless, these methods prioritize visual quality of the generated images over the preservation of utilities, for both annotated and unannotated useful attributes, undermining the data's usefulness for downstream ML tasks. \cite{hsu2020obfuscation} proposes to suppress sensitivity by only locating and obfuscating information-leaking features, but is also limited in providing a mechanism to quantify and preserve the utilities. To explicitly preserve useful annotated attributes,  ALR~\citep{bertran2019adversarially} ensures that annotated useful attributes remain predictable in anonymized data, while thwarting inference of sensitive attributes from an information-theoretic perspective. PPDAR~\citep{wu2020privacy} extends this approach by introducing a cross-entropy-based suppression and preservation loss. This idea is further blended with a prior-based suppression loss by BDQ~\citep{kumawat2022privacy}. Despite their advancements in preserving annotated useful attributes, these studies do not consider managing the unannotated attributes in the data.

\paragraph{Unannotated attributes management based on heuristics.} 
In the neighboring field of fair representation learning, ALFR~\citep{edwards2015censoring} proposes to preserve the unannotated attributes by minimizing the $\ell_2$ reconstruction loss, while selectively suppress and preserve annotated attributes. Building upon this, LAFTR~\citep{madras2018learning} introduces a fairness metrics driven optimization objective for suppression.
However, these studies focused on suppressing only one binary sensitive attribute to achieve fairness. In contrast, in IT Privacy literature, GAP~\citep{huang2018generative} advocates for suppressing multiple sensitive attribute, and simultaneously contraining the $\ell_2$ reconstruction loss. \cite{malekzadeh2019mobile} further combines $\ell_2$ reconstruction loss with information theoretic losses for annotated attributes. 
On the other hand, \citeauthor{dave2022spact} targets their work at suppressing the unannotated attributes of the data, utilizing contrastive learning technique, while ensuring the predictability of annotated attributes. Despite the practical relevance of their handling of unannotated attributes, these works fall short in providing a robust theoretical foundation regarding the derivation and operational bounds of their design, raising concerns in scenarios demanding high safety assurances. We discuss related works in more detail in Appendix \ref{app:rw}.

\section{Problem Formulation}

In this paper, we focus on a multi-attribute dataset comprised of original data $X$, a set of $M$ sensitive attributes $S = (S_1, S_2, \dots, S_M)$, a set of $N$ annotated useful attributes $U = (U_1, U_2, \dots, U_N)$, and a set of unannotated useful attributes or generic features $F$. However, our access is limited to the observable joint distribution $P(X, U, S)$, as opposed to the intrinsic joint distribution $P(X, U, S, F)$. We base our work on pragmatic assumptions that $S, U$ are random variables following finite categorical distributions, allowing the mutual information between $S, U$, and $X$ to be bounded. Additionally, we presuppose that with the given $X$, the corresponding annotated attributes $S, U$ are entirely determined (i.e., $P(S_i|X)$ and $P(U_j|X)$ are degenerate distributions). For broad applicability, we do not make assumptions regarding the dimension or distribution family for $F$ and $X$. Furthermore, we do not assume independence between $F$ and other variables, which means that $F$ may correlate with the joint distribution of $X, S, U$.

Our goal of IT Privacy is then formulated as finding the optimal data transformation $P_\theta(X'|X)$, where the random variable $X'$ is the transformed data, and the strongest unannotated useful attribute extractor $P_\eta(F|X)$ by solving the following constrained optimization problem:
\begin{equation}\label{eq:problem}
\begin{aligned}
 \max_{\theta, \eta} \quad & I(X';F)\\ 
 \text{s.t.} \quad & I(X' ; S_i) \leq m_i\;\text{and}\;I(X' ; U_j) \geq n_j,
\end{aligned}
\end{equation}
\begin{figure}[bt]
    \centering
    \includegraphics[width=0.3\linewidth]{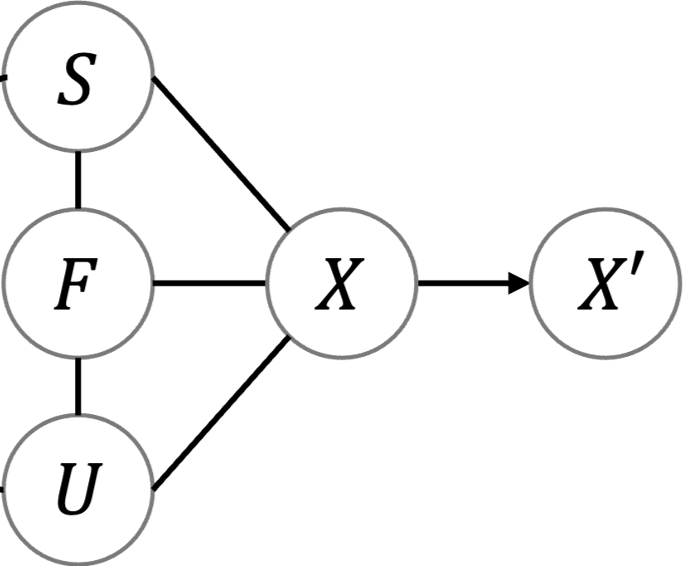}
    \vspace{-0.05in}
    \caption{\small The Markov chain of all variables. $F$ is correlated with $U,S,X$. $X'$ is only dependent on $X$.}
    \label{fig:mc}
    \vspace{-0.2in}
\end{figure}
where $I(\cdot,\cdot)$ is Shannon mutual information, $i \in 1\dots M$, $j \in 1\dots N$, $P_\theta(X'|X)$ and $P_\eta(F|X)$ are parameterized by $\theta, \eta$ respectively. By solving this optimization problem, we try to ensure that, at least $n_j$ nats (the counterpart of bits with Napierian base) information is preserved for $U_j$ in the transformed data $X'$, at most  $m_i$ nats information is leaked for $S_i$  in $X'$, and  the information preserved for $F$ in $X'$ is maximized when the most informative $F$ is extracted from $X$. For clarity, the Markov Chain of variables $U,S,F,X$, and $X'$ corresponding to our problem formulation is summarized in Figure \ref{fig:mc}.

\subsection{Operational Bounds}
In preparation for solving our optimization problem formulated in Equation \ref{eq:problem}, a thorough comprehension of its operational bounds is imperative. Specifically, we will elucidate formally that the parameters $m_i$ and $n_j$ must be chosen under certain constraints to ensure the solvability of Equation \ref{eq:problem}. Moreover, it will be established that the optimization objective $I(X';F)$ has an upperbound which can not be exceeded.
\begin{theorem}\label{th:mn}
For the Markov Chain shown in Figure \ref{fig:mc}, there exists a solution to the optimization problem defined in Equation \ref{eq:problem}, only if for any pair of $(m_i,n_j)$, $i \in 1\dots M$, $j \in 1\dots N$, it satisfies:
\begin{equation}\label{eq:mn}
     n_j \leq m_i + I(X;U_j|S_i), \quad   n_j \leq I(X;U_j)\; \textup{and} \;   m_i \geq 0.
\end{equation}
Under the assumptions that $P(S_i|X)$ and $P(U_j|X)$ are degenerate distributions, Equation \ref{eq:mn} can be simplified to
\begin{equation}\label{eq:mn2}
        n_j \leq m_i + H(U_j|S_i), \quad   n_j \leq H(U_j) \; \textup{and} \;    m_i \geq 0.
\end{equation}
where $H(\cdot)$ is Shannon entropy.

Besides, for any $m_i$, $i \in 1\dots M$, $I(X';F)$ is upper bounded by
\begin{equation}\label{eq:ob}
I(X';F) \leq H(X|S_i) + m_i.
\end{equation}
\end{theorem}
\vspace{-0.1in}
Please refer to Appendix~\ref{pf:mn} for the proof. It is important to note that the values in Equation \ref{eq:mn2}, specifically $H(U_j|S_i)$ and $H(U_j)$, are independent of our model's parameters and can be computed prior to training to assess solvability.

To understand the requirement of $n_j \leq m_i + H(U_j|S_i)$ in Equation \ref{eq:mn2} intuitively, consider a facial image dataset with two attributes ``hair color'' and ``age''. The high correlation between these attributes is evident, as older individuals are more likely to have white or gray hair.
Should ``age'' be suppressed with a small $m_{\text{age}}$, the ``hair color'' information in the facial image must be correspondingly sacrificed to prevent inadvertently disclosing ``age'' information.  The extent of this sacrifice is intuitively determined by the certainty with which ``age'' predicts ``hair color''.

To intuitively understand Equation \ref{eq:ob}, revert to the example we discussed above. When suppressing ``age'', certain features that were in $X$ no longer reside in $X'$, such as hair color and wrinkles, etc. This results in a necessary sacrifice of the information of $F$ contained in $X'$. The extent of sacrifice is determined by the certainty with which ``age'' determines the image.

\section{Data-driven Learnable Data Transformation Framework}\label{sec:method}
\begin{figure*}[tbp]
    \centering
    \includegraphics[width=.9\linewidth]{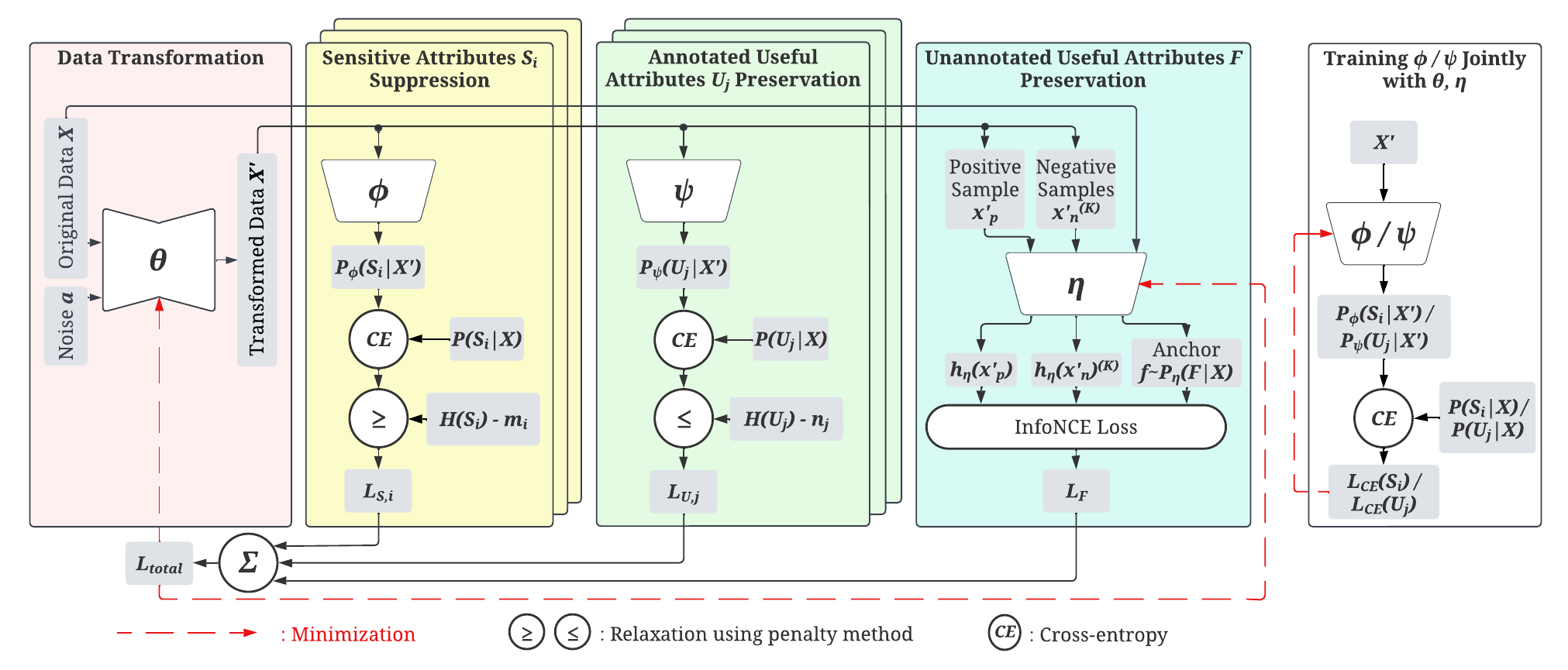}
    \caption{\small The overall architecture of MaSS. The data transformation module converts the original data into a transformed version. Then the transformed data is sent to both the sensitive attributes suppression module and the annotated useful attributes preservation module, to calculate a relaxed suppression or preservation loss for each attribute respectively. Additionally, the original and transformed data are sent to the unannotated useful attributes preservation module to calculate a contrastive loss. Finally, these losses are aggregated to minimize $\theta$ and $\eta$ jointly. $\phi,\psi$ are optimized with traditional supervised learning.}
    \label{fig:overall}
\end{figure*}

Building upon our problem formulation, we design a learnable data-driven data transformation framework as an approximation to Equation \ref{eq:problem}, which we call Multi-attribute Selective Suppression (abbreviated as MaSS).
Notably, we adopt neural networks as conditional probability approximators in our framework, and design our training objectives to be fully differentiable, allowing gradient descent based optimization. MaSS can be flexibly implemented with various neural network structures to adapt to different application requirements. The overarching architecture of MaSS is depicted in Figure \ref{fig:overall}. In the subsequent sections, we elaborate on the modules of MaSS in detail.

\subsection{Data Transformation}
The data transformation module takes in the original data $X$ and outputs the transformed data $X'$. In line with \citeauthor{bertran2019adversarially}, we parameterize $P_\theta(X'|X)$ as a neural network $X' = g_\theta(X, a)$, wherein $a$ is a noise variable sampled from a multi-variate unit Gaussian distribution, serving as the source of randomness for $X'$.

\subsection{Sensitive Attributes Suppression}\label{sec:s}

Given the transformed data, we calculate a suppression loss $L_{S,i}$ for each of the sensitive attributes, which is differentiable and can be minimized to achieve the constraint $I(X' ; S_i) \leq m_i$ mentioned in Equation \ref{eq:problem}. Next, we discuss in depth on the derivation of $L_{S,i}$.

The direct computation of $I(X' ; S_i)$ is infeasible because of the intractability of $P(S_i|X')$. Consequently, we incorporate an \textit{adversarial} neural network $P_\phii(S_i|X')$ as an estimation of $P(S_i|X')$, where $\phii$ is trained with the cross-entropy loss used in traditional supervised learning method:
 \begin{equation}\label{eq:phi}
 \begin{aligned}
      L_{CE}(S_i) &=  \E_{P(X)P_\theta(X'|X)} \left[H(P(S_i|X),P_\phii(S_i|X'))\right], \\
    \phii &= \arg\min_\phii L_{CE}(S_i),
 \end{aligned}
 \end{equation}
where $H(\cdot,\cdot)$ denotes cross-entropy and the expectation is estimated using mini-batch during training. Under the assumption that $S_i$ can be fully determined given $X$, $P(S_i|X)$ effectively refers to the deterministic ground-truth label of each sample.

With the help of $\phii$, the mutual information $I(X' ; S_i)$ can be estimated as
\begin{equation}
    \begin{aligned}
      I(X' ; S_i)   & \approx \E_{P(X)P_\theta(X'|X)P(S_i|X)} \left[\log \frac{P_\phii(S_i|X')}{P(S_i)}\right] \\
     & = H(S_i) - L_{CE}(S_i),
\end{aligned}
\end{equation}
where $H(S_i)$ is a constant for each dataset and can be calculated before training. Consequently, we can also convert the constraint $I(X' ; S_i) \leq m_i$ in Equation \ref{eq:problem} to
\begin{equation}\label{eq:suppress}
      H(S_i) - m_i   \leq L_{CE}(S_i).
\end{equation}
 Following \citeauthor{bertran2019adversarially}, we relax the Constraint \ref{eq:suppress} to a differentiable loss $L_{S,i}$ function eligible for gradient descent using the penalty method, which can be written as
\begin{equation}
\begin{aligned}
    d_{S,i} &= \min(L_{CE}(S_i) + m_i- H(S_i) , 0),\\
    L_{S,i} &=  d^{2}_{S,i} + |d_{S,i}|.
\end{aligned}
\end{equation}
\subsection{Annotated Useful Attributes Preservation}
The annotated useful attributes preservation module follows a symmetric design and derivation as the annotated sensitive attributes suppression module. Analogously, a differentiable preservation loss $L_{U,j}$ is calculated for each useful attribute to achieve the constraint $I(X' ; U_j) \geq n_j$ in Equation \ref{eq:problem}.  A \textit{collaborative} neural network $P_{\psij}(U_j|X')$ is also introduced to estimate $P(U_j|X')$, which is trained with cross-entropy loss:
\begin{equation}\label{eq:psi}
\begin{aligned}
    L_{CE}(U_j) &=\E_{P(X)P_\theta(X'|X)} [H(P(U_j|X), P_{\psij}(U_j|X'))], \\
            \psij &= \arg\min_{\psij} L_{CE}(U_j),
\end{aligned}
\end{equation}
Following the same derivation as Section \ref{sec:s}, we can convert the constraint $I(X' ; U_j) \geq n_j$ to
\begin{equation}\label{eq:preserve}
      H(U_j) - n_j   \geq L_{CE}(U_j).
\end{equation} 
which can also be relaxed into a differentiable loss $L_{U,j}$ using penalty method:
\begin{equation}
\begin{aligned}
    d_{U,j} &= \max(L_{CE}(U_j) + n_j - H(U_j), 0)\\
    L_{U,j}& =  d_{U,j}^2 + |d_{U,j}|.
\end{aligned}
\end{equation}
In order to accelerate the training process, we further propose to pre-train an attribute inference network on original data $X$ for each $S_i,U_i$, denoted as $\phiiz$ and $\psijz$ respectively, using the cross-entropy loss. And then we initialize the transformed data attribute inference models $\phii$ and $\psij$ with $\phiiz$ and $\psijz$ respectively, so that they can converge faster during training.

Note that, different from our method, previous studies such as \citeauthor{bertran2019adversarially} propose to freeze the useful attribute inference model $\psij$ during training after it is initialized with $\psijz$. However, we abandoned this strategy, because a frozen useful attribute inference model will introduce a noticeable error in estimating $I(X';U_j)$. Specifically, the estimation error will be $KL(P(U_j|X')||P_{\psijz}(U_j|X'))$, where $P_{\psijz}(U_j|X')$ denotes $P(U_j|X')$ estimated with the \textbf{frozen} useful attribute inference network $\psijz$, $KL(\cdot||\cdot)$ is Kullback–Leibler divergence. This error can be large and even unbounded because $\psijz$ is trained to approximate $P(U_j|X)$ rather than $P(U_j|X')$. Please refer to Appendix \ref{pf:preserve} for proof and analysis.

\subsection{Unannotated Useful Attributes Preservation}
The unannotated useful attributes preservation module aims at calculating and maximizing a differentiable loss approximating the negative $I(X';F)$, without any assumption on the distribution family of $F$ and $X'$. Consequently, we can not approximate $I(X';F)$ using the aforementioned method in Section \ref{sec:s} because it requires the assumption that $P(F|X')$ follows a finite categorical distribution.

Moreover, approximating $I(X';F)$ using $\ell_2$ reconstruction loss $ \left\| X' - F\right\|_2$ (or $ \left\| X' - X\right\|_2$), as did in \citet{huang2018generative}, \citet{malekzadeh2019mobile}, \citet{edwards2015censoring} and \citet{madras2018learning} is also infeasible since it also requires the assumptions that $P(F|X')$ (or $P(X|X')$) follows a fully factorized Gaussian distribution where each element of $F$ (or $X$) is only dependent on the corresponding element of $X'$ at the same position. In addition, we will also show empirically in Section \ref{sec:experiment} that $\ell_2$ reconstruction loss hinders the overall performance seriously due to its unrealistic assumptions.

In order to approximate $I(X';F)$ without assumption on the distribution family of $X'$ or $F$, we choose the InfoNCE loss $L_{F}$~\citep{oord2018representation} as the approximation of the shifted negative $I(X';F)$. InfoNCE is known as an effective approximation method for mutual information, regardless of the distribution family of the random variables, and is stable in mini-batch based training. We also tried other mutual information estimator, e.g., MINE~\citep{belghazi2018mine}, and empirically compared them in Section \ref{sec:experiment}.

To calculate $L_{F}$, we first sample one anchor $f$, one positive sample $x'_p$, and $K$ negative samples $x_n'^{(K)}$ given a specific realization of $X$, from the conditional distribution $P_\eta(F|X)$, $P_\theta(X'|X)$ and the marginal distribution $P_\theta(X')$ respectively. Then we take expectation over $X$ and all possible sampling of $f, x_p', x_n'^{(K)}$ to calculate $L_{F}$ as \begin{equation}\label{eq:contrast}
        L_{F} = \E \left[\log \frac{\mathcal{F}(f,x'_p)}{ \mathcal{F}(f,x'_p) + \sum_{x'_n \in x_n'^{(K)}} \mathcal{F}(f,x'_n) }\right],
\end{equation}
where $\mathcal{F}$ is a score function defined in the same way as SimCLR~\citep{chen2020simple}, which can be written as
\begin{equation}
    \mathcal{F}(f,x') = e^{\cos(f,h(x'))/\tau},
\end{equation}
where $\tau$ is the temperature hyper-parameter. $h(x')$ is a feature extractor trained jointly with data transformation module, $\theta$. 
Note that unlike SimCLR, our loss do not sample negative samples from $P(F)$. As proved in \citet{oord2018representation}, we can approximate $I(X';F) $ as
\begin{equation}
        I(X';F) \approx  - L_{F} +\log (K+1).
\end{equation}
In order to further encourage the transformed data $X'$ to remain in the original sample space of $X$, we propose to use a single neural network $\eta$ to parameterize both $h_\eta(x')$ and $P_\eta(F|X)$. This symmetric design can also reduce the number of parameters and hence stabilize the training. Importantly, an alternative interpretation of this design is to apply the InfoNCE loss on $X$ and $X'$ to estimate and maximize $I(X',X)$. 

Aligned with pretraining the attribute inference networks, our unannotated useful attributes extractor $\eta$ is also initialized with $\eta_0$ pretrained using InfoNCE loss on the original dataset $X$. In the pretraining stage we use one sample in the mini-batch as both the anchor and the positive sample and use the other samples in the mini-batch as negative samples.

Analogous to $L_{F}$, which is anchored in $F$ space,  we can define another InfoNCE loss $L'_{F}$ anchored in the $X'$ space and use both losses for training.  A more detailed elaboration on the calculation and the advantage of InfoNCE loss is presented in Appendix \ref{app:infonce}.

\subsection{Module Aggregation}
Aggregating the losses calculated from all modules above, we convert our original constrained optimization problem defined in Equation \ref{eq:problem} into the following differentiable  optimization problem:
\begin{equation}\label{eq:total}
    \min_{\theta,\eta} L\textsubscript{total} = \frac{L_F + L_F'}{2} + \lambda \left(\sum_i L_{S,i} + \sum_j L_{U,j}\right)
\end{equation}
where $\lambda$ is a hyper parameter controlling the degree of relaxation. Equation \ref{eq:total} effectively recovers the constrained optimization problem defined in Equation \ref{eq:problem} when $\lambda \to \infty$,.

\section{Evaluation}
In this section, we present our experimental evaluation of MaSS against several baselines methods using multiple datasets of varying modalities.

\subsection{Experimental Setup}\label{sec:experiment}
\textbf{Datasets.} The evaluation of MaSS is exhaustively conducted on three multi-attribute benchmark datasets of different modalities, namely the AudioMNIST~\citep{becker2018interpreting} dataset for recorded human voices, the Motion Sense~\citep{malekzadeh2019mobile} dataset for human activity sensor signals, and the Adience~\citep{eidinger2014age} dataset for facial images. We use the raw data points for training on Motion Sense and Adience, whereas we convert the raw data points to feature embeddings for AudioMNIST using state-of-the-art feature extractor HuBERT-B~\citep{hsu2021hubert} for training efficiency. 

\textbf{Baselines.} We compare our method with 5 baselines, namely ALR~\citep{bertran2019adversarially}, GAP~\citep{huang2018generative}, MSDA~\citep{malekzadeh2019mobile}, BDQ~\citep{kumawat2022privacy}, and PPDAR~\citep{wu2020privacy}. All 6 methods rely on adversarial training a sensitive attribute inference model. However, ALR, BDQ, and PPDAR do not consider the preservation of unannotated useful attributes, whereas GAP and MSDA do, using a $\ell_2$ heuristic loss. Notwithstanding, GAP does not consider the preservation of annotated useful attribute.

\textbf{Evaluation Metrics.} This paper is focused on suppressing sensitive attributes while preserving useful attributes, rather than generating high quality synthetic data. Therefore, we adopt classification accuracy for each attribute on evaluation set as our metric to measure the effectiveness of the suppression or preservation. Specifically, for sensitive attributes, we report the classification accuracy of the adversarially trained classifier $\phii$. For useful attributes, to ensure a fair comparison with baselines, we report the classification accuracy of a classifier tuned on the transformed data $X'$ and its attributes $U_j$. The performance is considered better when the sensitive attributes' accuracies are lower and the useful attributes' accuracies are higher. 

Furthermore, since the datasets we use are unbalanced, we adopt the classification accuracy of the majority classifier as a lower reference value, which can also be interpreted as the accuracy of \textit{guessing} the attribute without accessing $X'$ \citep{asoodeh2018estimation,liao2019tunable}. 
On the other hand, we also adopt the accuracy of the $\phiiz$ and $\psijz$ on original data $X$ in the evaluation set as a upper reference value of classification accuracy, which reflects the classification accuracy when no attributes are suppressed. 

Based on the lower and upper reference values of classification accuracy, we introduce a noval normalized metric for our task, namely Normalized Accuracy Gain (NAG), which is defined as $\mbox{NAG} = \max\left(0,\frac{Acc - Acc\textsubscript{guessing}}{Acc\textsubscript{no\_suppression} - Acc\textsubscript{guessing}}\right)$,
where $Acc$ denotes classification accuracy. NAG is inherently non-negative, with $\mbox{NAG}=0$ suggesting that $Acc\leq Acc\textsubscript{guessing}$. We consider all $Acc\leq Acc\textsubscript{guessing}$ as equally effective, which indicates that this attribute is completely suppressed from $X'$. NAG can be seen as a more informative indicator of how the classification accuracy of each attribute is increased or decreased. Therefore, we only report NAG throughout the main paper for clarity, while the corresponding results and reference values measured in classification accuracy are also shown in Appendix \ref{app:results}.%

In order to evaluate the performance of MaSS on preserving unannotated useful attributes, we conceal the labels (annotations) of certain annotated attributes during training and only use these labels for evaluation.

\textbf{Hyperparameters.} Throughout our experiments, $\lambda$ is simply set to 1. When an attribute is suppressed we simply set its mutual information constraint $m$ as 0. Unless otherwise noted, we set the $n$ of all preserved annotated attributes as the maximal value permitted by Equation~\ref{eq:mn2}. 

Additional detailed descriptions of the datasets, model structures, and optimization process are elaborated in Appendix \ref{app:experiment}.
Next we present and discuss our experimental results.

\begin{table}[!tb]
\caption{\small Comparison of the NAG between MaSS, ablations and baselines on Motion Sense. We suppress gender, ID, while preserve activity as if unannotated useful attribute.}
\label{tb:har1}
\begin{center}
\begin{adjustbox}{max width=0.85\linewidth}
\begin{tabular}{lccc}
\toprule
\multirow{2}{*}{Method} & \multicolumn{3}{c}{Normalized Accuracy Gain}                                                                         \\ \cmidrule(lr){2-4} 
                        & \multicolumn{1}{c}{gender ($\downarrow$)} & \multicolumn{1}{c}{ID ($\downarrow$)} & \multicolumn{1}{c}{activity ($\uparrow$)} \\ \midrule
ALR                     & \multicolumn{1}{c}{0.0828}       & \multicolumn{1}{c}{0.0432}   & 0.7704                            \\
GAP                     & \multicolumn{1}{c}{0.0053}       & \multicolumn{1}{c}{0.0314}   & 0.8379                            \\
MSDA                    & \multicolumn{1}{c}{0.0063}       & \multicolumn{1}{c}{0.0708}   & 0.8418                            \\
BDQ                     & \multicolumn{1}{c}{0.1178}       & \multicolumn{1}{c}{0.0613}   & 0.7426                            \\
PPDAR                   & \multicolumn{1}{c}{0.0000}       & \multicolumn{1}{c}{0.0000}   & 0.6912                            \\ \midrule
MaSS-NF                 & \multicolumn{1}{c}{0.0000}   & \multicolumn{1}{c}{0.0000}       & 0.7275       \\
MaSS-$\ell_2$           & \multicolumn{1}{c}{0.0085}   & \multicolumn{1}{c}{0.0260}       & 0.8156       \\
MaSS-MINE               & \multicolumn{1}{c}{0.3294}   & \multicolumn{1}{c}{0.1271}       & 0.6847       \\\midrule
MaSS                    & \multicolumn{1}{c}{0.0000}       & \multicolumn{1}{c}{0.0026}   & 0.8977                            \\
\bottomrule
\end{tabular}
\end{adjustbox}
\end{center}
\end{table}

\subsection{Evaluation on Human Activity Sensor Signals}
We first experiment on the human activity sensor signal dataset, Motion Sense. Initial experiment focuses on suppressing gender and ID attributes, while concealing the labels of the activity attribute, treating activity as an unannotated attribute for preservation. This setup mirrors scenarios aspiring to eliminate sensitive identity-related information from a dataset lacking explicit annotation on non-sensitive attributes. Apart from the 5 baselines described above, we also compare MaSS with 3 ablations to examine the unannotated useful attributes preservation module of MaSS: 1) removing the InfoNCE loss (denoted as MaSS-NF); 2) replacing InfoNCE loss to $\ell_2$ reconstruction loss (denoted as MaSS-$\ell_2$); and 3) replacing InfoNCE loss to a negative mutual information estimated using MINE~\citep{belghazi2018mine} (denoted as MaSS-MINE). Results as shown in Table \ref{tb:har1} demonstrate that MaSS attains the highest NAG on the activity attribute compared with all baselines and ablations. Additionally, in comparison to GAP, MSDA and MaSS-$\ell_2$, our method showcases a higher NAG on activity and a reduced NAG on both suppressed attributes. We believe it is because the unrealistic assumptions made by $\ell_2$ reconstruction loss overly restrict the flexibility of the data transformation. Moreover, MaSS outperforms MaSS-MINE in all attributes, which can be partly attributed to the instability of MINE mutual information estimator in the training process of our task. These results underscore MaSS's proficiency in maintaining a superior balance between preserving meaningful features and suppressing sensitive attributes.

We further experiment with suppressing gender, while preserving ID as annotated, and preserving activity as unannotated. Please refer to Appendix \ref{app:har} for results and corresponding analysis.

\begin{table}[!tb]
\caption{\small Comparison of the NAG between MaSS, ablations and baselines on AudioMNIST. We suppress gender, accent, age, ID, while preserve digit as if an unannotated attribute.}
\label{tb:audio1}
\begin{center}
\begin{adjustbox}{max width=\linewidth}
\begin{tabular}{lccccc}
\toprule
\multirow{2}{*}{Method} & \multicolumn{5}{c}{Normalized Accuracy Gain}                                                                                                                                        \\ \cmidrule(lr){2-6}
                        & \multicolumn{1}{c}{gender ($\downarrow$)} & \multicolumn{1}{c}{accent ($\downarrow$)} & \multicolumn{1}{c}{age ($\downarrow$)} & \multicolumn{1}{c}{ID ($\downarrow$)} & digit ($\uparrow$) \\ \midrule
ALR                     & \multicolumn{1}{c}{0.0000}       & \multicolumn{1}{c}{0.0000}       & \multicolumn{1}{c}{0.0000}    & \multicolumn{1}{c}{0.0004}   & 0.1036    \\
GAP                     & \multicolumn{1}{c}{0.0000}       & \multicolumn{1}{c}{0.0000}       & \multicolumn{1}{c}{0.0000}    & \multicolumn{1}{c}{0.0281}   & 0.9485    \\
MSDA                    & \multicolumn{1}{c}{0.0000}       & \multicolumn{1}{c}{0.0000}       & \multicolumn{1}{c}{0.0000}    & \multicolumn{1}{c}{0.0074}   & 0.9451    \\
BDQ                     & \multicolumn{1}{c}{0.0000}       & \multicolumn{1}{c}{0.0000}       & \multicolumn{1}{c}{0.0000}    & \multicolumn{1}{c}{0.0112}   & 0.5565    \\
PPDAR                   & \multicolumn{1}{c}{0.0000}       & \multicolumn{1}{c}{0.0000}       & \multicolumn{1}{c}{0.0000}    & \multicolumn{1}{c}{0.0016}   & 0.2839    \\ 
\midrule
MaSS-NF                 & \multicolumn{1}{c}{0.0000}       & \multicolumn{1}{c}{0.0001}       & \multicolumn{1}{c}{0.0000}    & \multicolumn{1}{c}{0.0000}   & 0.1846    \\
MaSS-$\ell_2$                & \multicolumn{1}{c}{0.0008}       & \multicolumn{1}{c}{0.0001}       & \multicolumn{1}{c}{0.0020}    & \multicolumn{1}{c}{0.0306}   & 0.9517    \\ 
MaSS-MINE                & \multicolumn{1}{c}{0.0076}       & \multicolumn{1}{c}{0.0000}       & \multicolumn{1}{c}{0.0112}    & \multicolumn{1}{c}{0.0434}   & 0.5031    \\ \midrule
MaSS                    & \multicolumn{1}{c}{0.0000}       & \multicolumn{1}{c}{0.0000}       & \multicolumn{1}{c}{0.0000}    & \multicolumn{1}{c}{0.0029}   & 0.9675    \\
\bottomrule
\end{tabular}
\end{adjustbox}
\end{center}
\end{table}

\subsection{Evaluation on Voice Audio Dataset}

Next, the application of MaSS is extended to the AudioMNIST dataset. The initial experiment involves the suppression of gender, accent, age, and ID attributes while treating digit as an unannotated attribute for preservation. Results of this experiment are shown in Table~\ref{tb:audio1}. We can observe that MaSS achieves the highest NAG on digit compared with all baselines and ablations, as well as a lower or equal NAG on suppressed attributes compared with GAP, MSDA and MaSS-$\ell_2$, further substantiating the limitation of the $\ell_2$ heuristic reconstruction loss and the effectiveness of MaSS.

In the subsequent experiment, we aim to suppress gender, accent, and age, while preserve digit as annotated and ID as unannotated. This scenario emulates conditions wherein the dataset encompasses both sensitive and useful annotated attributes, alongside with to-be-preserved unannotated attributes. It is observable from the results shown in Table \ref{tb:audio2} that MaSS secures the highest NAG on ID, along with a NAG on digit that is comparably high to other methods. Notably, although MSDA's NAG on ID is close to MaSS, it adversely bears higher NAG across all suppressed attributes.

We next conduct an ablation experiment of different configurations of suppressed and preserved attributes using MaSS on AudioMNIST. The configurations and their corresponding results are shown in Table~\ref{tb:audio3}. We can see that MaSS consistently achieves $\mbox{NAG}=0$ for most of the suppressed attributes, alongside with high NAG for preserved attributes. 

We also conducted ablation experiments of varying mutual information constraints $m$ and $n$, as well as an experiment comparing our method and SPAct~\citep{dave2022spact}. The results and analyses can be found in Appendix \ref{app:audio}.

\begin{table}[!tb]
\caption{\small Comparison of the NAG between MaSS and baselines on AudioMNIST. We suppress gender, accent, age, while preserve digit as annotated useful attribute, and preserve ID as if an unannotated attribute.}
\label{tb:audio2}
\begin{center}
\begin{adjustbox}{max width=\linewidth}
\begin{tabular}{lccccc}
\toprule
\multirow{2}{*}{Method} & \multicolumn{5}{c}{Normalized Accuracy Gain}                                                                                                                                      \\ \cmidrule(lr){2-6} 
                        & \multicolumn{1}{c}{gender ($\downarrow$)} & \multicolumn{1}{c}{accent ($\downarrow$)} & \multicolumn{1}{c}{age ($\downarrow$)} & \multicolumn{1}{c}{ID ($\uparrow$)} & digit ($\uparrow$) \\
\midrule
ALR                     & \multicolumn{1}{c}{0.0000}       & \multicolumn{1}{c}{0.0000}       & \multicolumn{1}{c}{0.0056}    & \multicolumn{1}{c}{0.7032} & 0.9994    \\
GAP                     & \multicolumn{1}{c}{0.0000}       & \multicolumn{1}{c}{0.0000}       & \multicolumn{1}{c}{0.0000}    & \multicolumn{1}{c}{0.7036} & 0.9579    \\
MSDA                    & \multicolumn{1}{c}{0.0015}       & \multicolumn{1}{c}{0.0013}       & \multicolumn{1}{c}{0.0323}    & \multicolumn{1}{c}{0.8428} & 0.9981    \\
BDQ                     & \multicolumn{1}{c}{0.0000}       & \multicolumn{1}{c}{0.0007}       & \multicolumn{1}{c}{0.0013}    & \multicolumn{1}{c}{0.4038} & 0.9980    \\
PPDAR                   & \multicolumn{1}{c}{0.0000}       & \multicolumn{1}{c}{0.0000}       & \multicolumn{1}{c}{0.0000}    & \multicolumn{1}{c}{0.7027} & 0.9983    \\ \midrule
MaSS                    & \multicolumn{1}{c}{0.0000}       & \multicolumn{1}{c}{0.0000}       & \multicolumn{1}{c}{0.0000}    & \multicolumn{1}{c}{0.8514} & 0.9983    \\
\bottomrule
\end{tabular}
\end{adjustbox}
\end{center}
\end{table}

\begin{table}[!tb]
\caption{\small Comparison of the NAG for different configurations of MaSS on AudioMNIST. $\checkmark$ denotes that this attribute is suppressed, while all other attributes are preserved as annotated useful attributes.}
\label{tb:audio3}
\begin{center}
\begin{adjustbox}{max width=\linewidth}
\begin{tabular}{l ccccc lllll}
\toprule
\multirow{2}{*}{Method} & \multicolumn{5}{c}{MaSS Suppressed Attributes}                                                                                                              & \multicolumn{5}{c}{Normalized Accuracy Gain}                                                                                                      \\ 
\cmidrule(lr){2-6} \cmidrule(lr){7-11} 
                        & \multicolumn{1}{c}{gender}       & \multicolumn{1}{c}{accent}       & \multicolumn{1}{c}{age}          & \multicolumn{1}{c}{ID}           & digit        & \multicolumn{1}{c}{gender} & \multicolumn{1}{c}{accent} & \multicolumn{1}{c}{age}    & \multicolumn{1}{c}{ID}     & \multicolumn{1}{c}{digit} \\ \midrule
\multirow{4}{*}{MaSS}   & \multicolumn{1}{c}{$\checkmark$} & \multicolumn{1}{c}{}             & \multicolumn{1}{c}{}             & \multicolumn{1}{c}{}             &              & \multicolumn{1}{c}{0.0000} & \multicolumn{1}{c}{0.9342} & \multicolumn{1}{c}{0.9574} & \multicolumn{1}{c}{0.9632} & 0.9972                     \\
                        & \multicolumn{1}{c}{$\checkmark$} & \multicolumn{1}{c}{$\checkmark$} & \multicolumn{1}{c}{}             & \multicolumn{1}{c}{}             &              & \multicolumn{1}{c}{0.0000} & \multicolumn{1}{c}{0.0000} & \multicolumn{1}{c}{0.9199} & \multicolumn{1}{c}{0.9372} & 0.9987                     \\
                        & \multicolumn{1}{c}{$\checkmark$} & \multicolumn{1}{c}{$\checkmark$} & \multicolumn{1}{c}{$\checkmark$} & \multicolumn{1}{c}{}             &              & \multicolumn{1}{c}{0.0000} & \multicolumn{1}{c}{0.0000} & \multicolumn{1}{c}{0.0000} & \multicolumn{1}{c}{0.8680} & 0.9964                     \\
                        & \multicolumn{1}{c}{$\checkmark$} & \multicolumn{1}{c}{$\checkmark$} & \multicolumn{1}{c}{$\checkmark$} & \multicolumn{1}{c}{$\checkmark$} &              & \multicolumn{1}{c}{0.0000} & \multicolumn{1}{c}{0.0000} & \multicolumn{1}{c}{0.0000} & \multicolumn{1}{c}{0.0017} & 0.9953                     \\ \bottomrule
                        
\end{tabular}
\end{adjustbox}
\end{center}
\end{table}

\subsection{Evaluation on Facial Images}

Finally, we apply MaSS to Adience, suppressing gender while treating age and ID as unannotated attributes that should be preserved. The results shown in Table \ref{tb:face1} reveal that, among all methods with $\mbox{NAG}=0$ for gender, MaSS accomplishes the highest NAG for the preserved attributes. 

Additionally, we also empirically show that the transformed facial images can be accurately exploited by off-the-shelf pre-trained landmark detection model PIPNet~\cite{jin2021pixel}. The NME (Normalized Mean Error)~\citep{jin2021pixel} of PIPNet between transformed Adience and original Adience is 3.30\%, in comparison with the 3.94\% NME of PIPNet between original WLFW dataset~\citep{wu2018look} and ground truth label. The comparable performance showed that transformed Adience dataset can be accurately exploited by pre-trained PIPNet. %

Visualized transformed images, together with additional results on suppressing age, and an ablation study on retraining the sensitive attribute inference model are shown in Appendix \ref{app:face}.

Apart from the above mentioned 3 datasets, we additionally experiment MaSS on a tabular dataset~\citep{marketing} . The experimental setup, results and analysis are shown in Appendix~\ref{app:tabular}, which similarly validate the generalizability and effectiveness of MaSS.

\begin{table}[!tb]
\caption{\small Comparison of the NAG between MaSS and baselines on Adience. We suppress gender, while preserve age, ID as if unannotated useful attributes.}
\label{tb:face1}
\begin{center}
\begin{adjustbox}{max width=0.65\linewidth}
\begin{tabular}{lccc}
\toprule
\multirow{2}{*}{Method} & \multicolumn{3}{c}{Normalized Accuracy Gain}                                                                   \\ \cmidrule(lr){2-4} 
                        & \multicolumn{1}{c}{gender ($\downarrow$)} & \multicolumn{1}{c}{age ($\uparrow$)}  & \multicolumn{1}{c}{ID ($\uparrow$)} \\ \midrule
ALR                     & \multicolumn{1}{c}{0.0128}       & \multicolumn{1}{c}{0.0023}   & 0.0128                      \\
GAP                     & \multicolumn{1}{c}{0.0000}       & \multicolumn{1}{c}{0.4907}   & 0.5616                      \\
MSDA                    & \multicolumn{1}{c}{0.3114}       & \multicolumn{1}{c}{0.7928}   & 0.8461                      \\
BDQ                     & \multicolumn{1}{c}{0.0026}       & \multicolumn{1}{c}{0.0000}   & 0.0075                      \\
PPDAR                   & \multicolumn{1}{c}{0.0000}       & \multicolumn{1}{c}{0.0000}   & 0.0000                      \\ \midrule
MaSS                    & \multicolumn{1}{c}{0.0000}       & \multicolumn{1}{c}{0.7418}   &     0.7662                 \\\bottomrule
\end{tabular}
\end{adjustbox}
\end{center}
\end{table}

\section{Conclusion}
In this paper, we present MaSS, a generalizable and highly configurable 
data-driven learnable data transformation framework
that is capable of suppressing sensitive/private information from data while preserving its utility.
Compared to existing privacy protection techniques that have similar objectives,
MaSS is superior by satisfying all 5 desired properties of SUIFT.
We thoroughly evaluated MaSS on three datasets of different modalities, namely
voice recordings, human activity motion sensor signals, and facial images, 
and obtained promising results that demonstrate MaSS' practical effectiveness
under various tasks and configurations.

\section*{Impact Statement}
We believe that there is no ethical concern or negative societal consequence related to this work. Our work benefits the protection of people's privacy in that it is proposed to suppress sensitive attributes in the datasets while preserving their potential utility for downstream tasks. 

\section*{Disclaimer}
This paper was prepared for informational purposes by the Global Technology Applied Research center of JPMorgan Chase \& Co. This paper is not a product of the Research Department of JPMorgan Chase \& Co. or its affiliates. Neither JPMorgan Chase \& Co. nor any of its affiliates makes any explicit or implied representation or warranty and none of them accept any liability in connection with this paper, including, without limitation, with respect to the completeness, accuracy, or reliability of the information contained herein and the potential legal, compliance, tax, or accounting effects thereof. This document is not intended as investment research or investment advice, or as a recommendation, offer, or solicitation for the purchase or sale of any security, financial instrument, financial product or service, or to be used in any way for evaluating the merits of participating in any transaction.

\bibliography{icml2024}
\bibliographystyle{icml2024}

\newpage
\appendix
\onecolumn

\section*{Appendix}
\section{Proofs}\label{app:proofs}
\subsection{Proof of Theorem \ref{th:mn}}
\begin{proof}\label{pf:mn}
\textbf{Proof for Equation \ref{eq:mn}.} For the Markov Chain shown in Figure \ref{fig:mc}, for any $i \in 1\dots M$, and $j \in 1\dots N$, if both $I(X' ; S_i) \leq m_i$ and $I(X' ; U_j) \geq n_j$ hold, then we have
\begin{equation}
    \begin{aligned}
            m_i + I(X;U_j|S_i) &\geq I(X'; S_i)+ I(X;U_j|S_i)  \\
            &=I(X'; S_i)+ I(X',X;U_j|S_i)\\
            &= I(X'; U_j, S_i) -I(X'; U_j|S_i)+ I(X',X;U_j|S_i) \\
            &=I(X'; U_j, S_i) + I(X;U_j|X',S_i) \\
            &=I(X'; U_j) +I(X'; S_i| U_j)+ I(X;U_j|X',S_i) \\
            &\geq I(X'; U_j)\\
            &\geq n_j
    \end{aligned}
\end{equation}
which proves the first inequation.
 Following Data Processing Inequality, we can also have
\begin{equation}
    n_j \leq I(X' ; U_j)\leq I(X;U_j)
\end{equation}
which proves the second inequation. Finally, we also have
\begin{equation}
    m_i \geq I(X' ; S_i)\geq 0
\end{equation}
which proves the third inequation.

\textbf{Proof for Equation \ref{eq:mn2}.} 
Under the assumption that $U, S$ are fully determined given $X$ ($P(S_i|X),P(U_j|X)$ are degenerate distributions), we can have
\begin{equation}
    H(S_i|X) = 0, \qquad H(U_j|X)=0
\end{equation}
for any $i \in 1\dots M$, and $j \in 1\dots N$. Since adding a condition can not increase the entropy, we can also have
\begin{equation}
    0\leq H(U_j|X,S_i) \leq H(U_j|X)=0
\end{equation}
Therefore we have
\begin{equation}
    H(U_j|X,S_i)=0
\end{equation}

Inserting $H(U_j|X,S_i)=0$ and $H(U_j|X)=0$ into  the inequations \ref{eq:mn}, we can further convert them to the inequations \ref{eq:mn2} as
\begin{equation}
    \begin{aligned}
        n_j &\leq m_i + I(X;U_j|S_i)  \\
       & =  m_i + H(U_j|S_i) -  H(U_j|X,S_i) \\
        &=  m_i + H(U_j|S_i),
    \end{aligned}
\end{equation}
and
\begin{equation}
    \begin{aligned}
        n_j &\leq I(X;U_j)  \\
       & =  H(U_j) -  H(U_j|X) \\
        &= H(U_j).
    \end{aligned}
\end{equation}

\textbf{Proof for Equation \ref{eq:ob}.} 
For the Markov Chain shown in Figure \ref{fig:mc}, according to Data Processing Inequality, we have 
    \begin{equation}
            I(X';X) - I(X';F) = I(X';X|F) \geq 0
    \end{equation}
Therefore, we have
    \begin{equation}
           I(X';F)  \leq I(X';X)
    \end{equation}
We can also have
\begin{equation}
\begin{aligned}
        I(X';X) & = H(X') - H(X'|X) \\
        &= H(X') - H(X'|X,S_i) \\
        &\leq H(X') - H(X'|X,S_i) + H(X|X',S_i) \\
        &= H(X') +H(X|S_i) - H(X'|S_i) \\
        &= I(X';S_i) +H(X|S_i) \\
        &\leq H(X|S_i) + m_i
\end{aligned}
\end{equation}

\end{proof}

\subsection{Proof and Analysis for the Estimation Error of $I(X';U_j)$ with Frozen Useful Attribute Inference Network}\label{pf:preserve}

Let $P_{\psijz}(U_j|X')$ and $I_{\psijz}(X';U_j)$ denote the conditional distribution of $U_j$ given $X'$ and the mutual information between $U_j$ and $X'$ estimated with the \textbf{frozen} useful attribute inference network $\psijz$. For the Markov Chain shown in Figure \ref{fig:mc}, $I_{\psijz}(X';U_j)$ is calculated as
\begin{equation}
    I_{\psijz}(X';U_j) =  \E_{P(X)P_\theta(X'|X)P(U_j|X)} [\log \frac{P_{\psijz}(U_j|X')}{P(U_j)}]
\end{equation}
Therefore, we can prove
\begin{equation}
\begin{aligned}
         I(X';U_j) - I_{\psijz}(X';U_j) &= \E_{P(X)P_\theta(X'|X)P(U_j|X)} [\log\frac{P(U_j|X')}{P_{\psijz}(U_j|X')}] \\
         &= \E_{P_\theta(X')P(U_j|X')} [\log\frac{P(U_j|X')}{P_{\psijz}(U_j|X')}] \\
         &= KL(P(U_j|X')||P_{\psijz}(U_j|X'))
\end{aligned}
\end{equation}
where $KL(P(U_j|X')||P_{\psijz}(U_j|X'))$ can be large and even unbounded, because $\psijz$ is trained to approximate $P(U_j|X)$ rather than $P(U_j|X')$. Therefore, this strategy is not adopted in our design. 

\section{Additional Descriptions of Related Works}\label{app:rw}

In this section we present additional discussions on related works, especially on why they are categorized as not satisfying or partially satisfying certain properties of SUIFT. DeepPrivacy~\citep{hukkelaas2019deepprivacy} and CiaGAN~\citep{maximov2020ciagan} are considered partially satisfying U and F because the CGAN based framework would prioritize the visual quality of generated samples (whether they are differentiable by discriminator) over the preservation of useful information (whether generated samples contain the same useful information as their original prototypes). Besides, they do not provide theoretical justification for the CGAN based framework. \cite{hsu2020obfuscation} is considered partially satisfying F because it does not enforce the preservation of generic features explicitly. On the contrary it proposes to only obfuscate information-leaking features, while keep other features unaltered, which implicitly preserves the generic features. In addition, \cite{hsu2020obfuscation} also does not consider the existence of useful annotated attributes. ALR~\citep{bertran2019adversarially} proposes a rigorous information-theoretic framework for annotated attributes, but does not incorporate unannotated attributes in the discussion. PPDAR~\citep{wu2020privacy} and BDQ~\citep{kumawat2022privacy} also propose effective solutions for annotated attributes preservation or suppression. But they do not take unannotated attributes and a detailed theoretical foundation into consideration as well. 

For the works that consider the management of unannotated attributes, ALFR~\citep{edwards2015censoring} and LAFTR~\citep{madras2018learning} are proposed in fairness literature, which are designed to release a compact representation for downstream tasks, and only suppress a binary sensitive attribute. GAP~\citep{huang2018generative} does not consider the preservation of annotated useful attributes. And importantly, all of ALFR, LAFTR, GAP and MSDA~\citep{malekzadeh2019mobile} do not theoretically justify their design for unannotated attributes in the same framework proposed for annotated attributes. SPAct~\citep{dave2022spact} does not consider the existence of annotated senstive attributes and does not theoretically justify its design.

\section{Additional Description of Proposed Method}
\subsection{InfoNCE Contrastive Learning}\label{app:infonce}
InfoNCE contrastive learning loss~\cite{oord2018representation} is a classical contrastive learning loss, which learns useful representations of data by making the representations of positive samples (similar or related samples) closer while pushing the representations of negative samples further apart from the anchor. The sampling strategy in our framework is as follows. Suppose we have ${K+1}$ samples $ \{ x _i \}  _{i=1}^{K+1}$ in a mini-batch. We first pass them through the feature extractor $P _\eta(F|X)$ and data transformation module to sample a batch of $ \{ f _i \}  _{i=1}^{K+1}$ and $ \{ x' _i \}  _{i=1}^{K+1}$ respectively. Then suppose we choose the $j$-th feature $f _j$ as the anchor. Then the corresponding $x' _j$ would be designated as positive sample, and all other $x' _{i \neq j}$ are  designated as negative samples. After sampling, we calculate the contrastive learning loss as Equation 14 in our paper. For training stability, in our implementation each of $K+1$ features in a batch is used as anchor once and then averaged. 

An analogous InfoNCE contrastive learning loss $L'_{F}$ is anchored in $X'$ space, which is defined as 
\begin{equation}
   L'_{F} :=  \E_{x \sim P(X)} \E_{x' \sim P_\theta(X'|X)} \E_{{f_p \sim P_\eta(F|X)}} \E_{f_n^{(K)} \sim P_\eta(F) }   [\log \frac{\mathcal{F}(f_p,x')}{ \mathcal{F}(f_p,x') + \sum_{f_n \in f_n^{(K)}} \mathcal{F}(f_n,x') }]
\end{equation}
where $x',f_p,f_n^{(K)}$ are the anchor, the positive sample, and the negative samples respectively.

Compared with $\ell_2$ reconstruction loss, our contrastive learning loss is advantageous in that it does not presuppose the distributions of $F, X'$, making it broadly applicable across various domains like images, language, and sensor signals. Moreover, its superior empirical effectiveness is demonstrated in our experiments.

\section{Additional Experimental Setup}\label{app:experiment}
\subsection{Datasets}

We next introduce the datasets. The Adience dataset, consisting of 26580 facial images, was originally published to help study the recognition of age and gender. Each face image has 3 attributes: ID, gender and age. We filter out the IDs with only one image. For the rest of data points, we split them into training and evaluation set as 7:3, and ensure that for each ID there is at least one image in training set and one image in evaluation set. Data points used in our experiment contains 1042 different DataIDs, 8 age groups, and 2 gender classes. The images are resized to 80*80, converted to grayscale images, and normalzied to 0-1 in our experiments.

The AudioMNIST dataset contains audio recordings of spoken digits (0-9) in English from 60 speakers. The dataset contains 8 attributes, from which we used 5 most representative attributes for our experiments, namely gender, accent, age, ID, spoken digits, with 2, 16, 18, 60, 10 classes, respectively. There are 30,000 audio clips in total. We split the data into 24000, 6,000 for training and evaluation. The audio data are transformed to feature embeddings by HuBERT-B feature extractor and normalized to unit L2-norm.

The Motion Sense dataset contains the accelerometer and gyroscope data for human doing 6 daily activities. It contains 5 attributes, form which we used 3 most representative attributes for our experiments, namely gender, ID, and activity, with 2, 24, 6 classes respectively. Following \cite{malekzadeh2019mobile}, we did not use "sit" and "stand up" activity in experiments. We used the same split and data pre-processing method as \cite{malekzadeh2019mobile}, which resulted in 74324 segmented data points. Specifically, we used "trail" split strategy as described in \cite{malekzadeh2019mobile}, and we only used the magnitude of gyroscope and accelerometer as input. Signals are normalized to 0-mean and 1-std, and then cut into 128-length clips.

\subsection{Model Structures and Optimization}\label{app:model}

We elaborated the model structures and optimization methods used for our experiments in Table \ref{tb:model}. For faster convergence and training stability, we design the $\phi,\psi, \eta$ models used in facial image experiments as a fixed FaceNet~\citep{schroff2015facenet} backbone followed by learnable 3-layer MLPs, and design the $\theta$ model of facial image experiment as U-Net~\citep{ronneberger2015u}. For the same reason, we add residual structures from input of the first layer to the output of the second layer for 3-layer MLP $\theta$ models used in audio and human activity experiments.

\begin{table}[!tb]
\caption{Model structures and optimization methods used for our experiments.}
\label{tb:model}
\begin{center}
\begin{adjustbox}{max width=\linewidth}
\begin{tabular}{|l|ccc|}
\hline
Experiment                        & \multicolumn{1}{c|}{Audio}       & \multicolumn{1}{c|}{Human activity} & Facial image                                                                                          \\ \hline
Dataset                           & \multicolumn{1}{c|}{AudioMNIST}  & \multicolumn{1}{c|}{Motion Sense}                 & Adience                                                                                               \\ \hline
\# total data points         & \multicolumn{1}{c|}{30000}         & \multicolumn{1}{c|}{74324}                          & 26580                                                                                                  \\ \hline
Training-evaluation split         & \multicolumn{1}{c|}{4:1}         & \multicolumn{1}{c|}{7:4}                          & 7:3                                                                                                  \\ \hline
Optimizer                         & \multicolumn{3}{c|}{AdamW~\citep{loshchilov2017decoupled}}                                                                                                                                                                   \\ \hline
Learning rate                     & \multicolumn{3}{c|}{0.0001}                                                                                                                                                                  \\ \hline
Weight decay                      & \multicolumn{3}{c|}{0.001}                                                                                                                                                                   \\ \hline
Learning rate scheduler           & \multicolumn{3}{c|}{Cosine scheduler}                                                                                                                                                        \\ \hline
Epoch                             & \multicolumn{1}{c|}{2000}        & \multicolumn{1}{c|}{200}                          & 4000                                                                                                  \\ \hline
$\theta$ model structure          & \multicolumn{1}{c|}{3-layer MLP} & \multicolumn{1}{c|}{3-layer MLP}                  & U-Net                                                                                                 \\ \hline
$\phi,\psi, \eta$ model structure & \multicolumn{1}{c|}{3-layer MLP} & \multicolumn{1}{c|}{6-layer Convolutional NN}     & \begin{tabular}[c]{@{}c@{}}Fixed FaceNet backbone\\  followed by\\ learnable 3-layer MLP\end{tabular} \\ \hline
\end{tabular}
\end{adjustbox}
\end{center}
\end{table}

\section{Additional Experiments Results}\label{app:results}
\subsection{Evaluation on Human Activity Sensor Signals}\label{app:har}

In addition to the experimental results measured in NAG shown in Table \ref{tb:har1}, we also show the experimental results measured in accuracy below in Table \ref{tb:har1acc}.

We also conducted an experiment where we suppress gender, while preserve ID as annotated attribute, and preserve activity as unannotated attribute. We set the $n$ for ID as 1.6, which meets the requirements of Equation~\ref{eq:mn2}. The results are shown in Table~\ref{tb:har2}. We can observe that MaSS achieved lowest NAG on gender as well as comparable NAG on the other preserved attributes.  This outcome stems from the fact the sensitive attribute gender is determined by ID, therefore when we suppress gender, the information retained for ID is inherently limited as Equation~\ref{eq:mn2}. MaSS is explicitly aware of this limit and is adjusted to preserve only limited amount of information for ID. In contrast other baselines can only heuristically trade-off between suppressing and preserving.

\begin{table}[!tb]
\centering
\caption{\small Comparison of the accuracy and NAG between MaSS, ablations and baselines on Motion Sense. We suppress gender, ID, while preserve activity as if unannotated useful attribute.}
\vspace{0.1in}
\label{tb:har1acc}
\begin{adjustbox}{max width=\linewidth}
\begin{tabular}{llll}
\toprule
\multirow{2}{*}{Method} & \multicolumn{3}{c}{Accuracy (Normalized Accuracy Gain)}                                                                         \\ \cmidrule(lr){2-4} 
                        & \multicolumn{1}{c}{gender ($\downarrow$)} & \multicolumn{1}{c}{ID ($\downarrow$)} & \multicolumn{1}{c}{activity ($\uparrow$)} \\ \midrule
No suppression          & \multicolumn{1}{l}{0.9817 (1.0000)}       & \multicolumn{1}{l}{0.9026 (1.0000)}   & 0.9764 (1.0000)                            \\
Guessing                & \multicolumn{1}{l}{0.5699 (0.0000)}       & \multicolumn{1}{l}{0.0533 (0.0000)}   & 0.4663 (0.0000)                            \\ \midrule
ALR                     & \multicolumn{1}{l}{0.6040 (0.0828)}       & \multicolumn{1}{l}{0.0900 (0.0432)}   & 0.8593 (0.7704)                            \\
GAP                     & \multicolumn{1}{l}{0.5721 (0.0053)}       & \multicolumn{1}{l}{0.0800 (0.0314)}   & 0.8937 (0.8379)                            \\
MSDA                    & \multicolumn{1}{l}{0.5725 (0.0063)}       & \multicolumn{1}{l}{0.1134 (0.0708)}   & 0.8957 (0.8418)                            \\
BDQ                     & \multicolumn{1}{l}{0.6184 (0.1178)}       & \multicolumn{1}{l}{0.1054 (0.0613)}   & 0.8451 (0.7426)                            \\
PPDAR                   & \multicolumn{1}{l}{0.5698 (0.0000)}       & \multicolumn{1}{l}{0.0498 (0.0000)}   & 0.8189 (0.6912)                            \\ \midrule
MaSS-NF                 & \multicolumn{1}{c}{0.5699 (0.0000)}   & \multicolumn{1}{c}{0.0508 (0.0000)}       & 0.8374 (0.7275)       \\
MaSS-$\ell_2$                & \multicolumn{1}{c}{0.5734 (0.0085)}   & \multicolumn{1}{c}{0.0754 (0.0260)}       & 0.8823 (0.8156)       \\
MaSS-MINE               & \multicolumn{1}{c}{0.7056 (0.3294)}   & \multicolumn{1}{c}{0.1613 (0.1271)}       & 0.8156 (0.6847)       \\\midrule
MaSS                    & \multicolumn{1}{l}{0.5686 (0.0000)}       & \multicolumn{1}{l}{0.0555 (0.0026)}   & 0.9242 (0.8977)                           \\
\bottomrule
\end{tabular}
\end{adjustbox}

\end{table}

\begin{table}[!tb]
\caption{Comparison of the classification accuracy and NAG between MaSS and baselines on Motion Sense. We suppress gender, while preserve ID as annotated useful attribute, and preserve activity as if an unannotated attribute.}
\label{tb:har2}
\begin{center}
\begin{adjustbox}{max width=0.7\linewidth}
\begin{tabular}{lccc}
\toprule
\multirow{2}{*}{Method} & \multicolumn{3}{c}{Accuracy (Normalized Accuracy Gain)}                                                  \\ \cmidrule(lr){2-4} 
                        & \multicolumn{1}{c}{gender ($\downarrow$)} & \multicolumn{1}{c}{ID ($\uparrow$)} & activity ($\uparrow$) \\ \midrule
No suppression          & \multicolumn{1}{c}{0.9817 (1.0000)}       & \multicolumn{1}{c}{0.9026 (1.0000)} & 0.9764 (1.0000)       \\
Guessing                & \multicolumn{1}{c}{0.5699 (0.0000)}       & \multicolumn{1}{c}{0.0533 (0.0000)} & 0.4663 (0.0000)       \\ \midrule
ALR                     & \multicolumn{1}{c}{0.8258 (0.6214)}       & \multicolumn{1}{c}{0.6147 (0.6610)} & 0.8966 (0.8436)       \\
GAP                     & \multicolumn{1}{c}{0.6599 (0.2186)}       & \multicolumn{1}{c}{0.6628 (0.7176)} & 0.9378 (0.9243)       \\
MSDA                    & \multicolumn{1}{c}{0.6418 (0.1746)}       & \multicolumn{1}{c}{0.6360 (0.6861)} & 0.9030 (0.8561)       \\
BDQ                     & \multicolumn{1}{c}{0.7092 (0.3383)}       & \multicolumn{1}{c}{0.6583 (0.7124)} & 0.9269 (0.9030)       \\
PPDAR                   & \multicolumn{1}{c}{0.7830 (0.5175)}       & \multicolumn{1}{c}{0.5680 (0.6060)} & 0.8867 (0.8242)       \\ \midrule
MaSS                    & \multicolumn{1}{c}{0.5870 (0.0415)}       & \multicolumn{1}{c}{0.5931 (0.6356)} & 0.9168 (0.8832)       \\
\bottomrule
\end{tabular}
\end{adjustbox}
\end{center}
\end{table}

\subsection{Evaluation on Voice Audio Dataset}\label{app:audio}

In addition to the experimental results measured in NAG shown in Table \ref{tb:audio1} and Table \ref{tb:audio2}, we also show the experimental results measured in accuracy below in Table \ref{tb:audio1acc} and Table \ref{tb:audio2acc} respectively.

We also compare our method with SPAct \cite{dave2022spact}. Since SPAct does not consider preserving unannotated useful attributes. Therefore we compare it in a scenerio where we only have annotated attributes. We can observe that MaSS achieved slightly lower NAG on digit compared with SPAct, but significantly lower NAG on all sensitive attributes (up to 5\%), which shows that our method may achieve a better trade-off between suppression and preservation.

\begin{table}[!tb]
\caption{\small Comparison of the classification accuracy and NAG between MaSS, ablations and baselines on AudioMNIST. We suppress gender, accent, age, ID, while preserve digit as if an unannotated attribute.}
\label{tb:audio1acc}
\begin{center}
\begin{adjustbox}{max width=1\linewidth}
\begin{tabular}{lccccc}
\toprule
\multirow{2}{*}{Method} & \multicolumn{5}{c}{Accuracy (Normalized Accuracy Gain)}                                                                                                                                        \\ \cmidrule(lr){2-6}
                        & \multicolumn{1}{c}{gender ($\downarrow$)} & \multicolumn{1}{c}{accent ($\downarrow$)} & \multicolumn{1}{c}{age ($\downarrow$)} & \multicolumn{1}{c}{ID ($\downarrow$)} & digit ($\uparrow$) \\ \midrule
No suppression          & \multicolumn{1}{c}{0.9962 (1.0000)}       & \multicolumn{1}{c}{0.9843 (1.0000)}       & \multicolumn{1}{c}{0.9657 (1.0000)}    & \multicolumn{1}{c}{0.9808 (1.0000)}   & 0.9975 (1.0000)    \\
Guessing                & \multicolumn{1}{c}{0.8000 (0.0000)}       & \multicolumn{1}{c}{0.6833 (0.0000)}       & \multicolumn{1}{c}{0.1667 (0.0000)}    & \multicolumn{1}{c}{0.0167 (0.0000)}   & 0.1000 (0.0000)    \\ \midrule
ALR                     & \multicolumn{1}{c}{0.8000 (0.0000)}       & \multicolumn{1}{c}{0.6833 (0.0000)}       & \multicolumn{1}{c}{0.1667 (0.0000)}    & \multicolumn{1}{c}{0.0171 (0.0004)}   & 0.1930 (0.1036)    \\
GAP                     & \multicolumn{1}{c}{0.8000 (0.0000)}       & \multicolumn{1}{c}{0.6828 (0.0000)}       & \multicolumn{1}{c}{0.1663 (0.0000)}    & \multicolumn{1}{c}{0.0438 (0.0281)}   & 0.9513 (0.9485)    \\
MSDA                    & \multicolumn{1}{c}{0.8000 (0.0000)}       & \multicolumn{1}{c}{0.6833 (0.0000)}       & \multicolumn{1}{c}{0.1665 (0.0000)}    & \multicolumn{1}{c}{0.0238 (0.0074)}   & 0.9482 (0.9451)    \\
BDQ                     & \multicolumn{1}{c}{0.8000 (0.0000)}       & \multicolumn{1}{c}{0.6833 (0.0000)}       & \multicolumn{1}{c}{0.1667 (0.0000)}    & \multicolumn{1}{c}{0.0275 (0.0112)}   & 0.5995 (0.5565)    \\
PPDAR                   & \multicolumn{1}{c}{0.8000 (0.0000)}       & \multicolumn{1}{c}{0.6833 (0.0000)}       & \multicolumn{1}{c}{0.1667 (0.0000)}    & \multicolumn{1}{c}{0.0182 (0.0016)}   & 0.3548 (0.2839)    \\ 
\midrule
MaSS-NF                 & \multicolumn{1}{c}{0.8000 (0.0000)}       & \multicolumn{1}{c}{0.6833 (0.0001)}       & \multicolumn{1}{c}{0.1658 (0.0000)}    & \multicolumn{1}{c}{0.0152 (0.0000)}   & 0.2657 (0.1846)    \\
MaSS-$\ell_2$                & \multicolumn{1}{c}{0.8002 (0.0008)}       & \multicolumn{1}{c}{0.6833 (0.0001)}       & \multicolumn{1}{c}{0.1683 (0.0020)}    & \multicolumn{1}{c}{0.0462 (0.0306)}   & 0.9542 (0.9517)    \\ 
MaSS-MINE                & \multicolumn{1}{c}{0.8015 (0.0076)}       & \multicolumn{1}{c}{0.6833 (0.0000)}       & \multicolumn{1}{c}{0.1757 (0.0112)}    & \multicolumn{1}{c}{0.0585 (0.0434)}   & 0.5515 (0.5031)    \\ \midrule
MaSS                    & \multicolumn{1}{c}{0.8000 (0.0000)}       & \multicolumn{1}{c}{0.6833 (0.0000)}       & \multicolumn{1}{c}{0.1667 (0.0000)}    & \multicolumn{1}{c}{0.0195 (0.0029)}   &0.9683 (0.9675)    \\
\bottomrule
\end{tabular}
\end{adjustbox}
\end{center}
\end{table}

\begin{table}[!tb]
\caption{\small Comparison of the classification accuracy and NAG between MaSS and baselines on AudioMNIST. We suppress gender, accent, age, while preserve digit as annotated useful attribute, and preserve ID as if an unannotated attribute.}
\label{tb:audio2acc}
\begin{center}
\begin{adjustbox}{max width=\linewidth}
\begin{tabular}{lccccc}
\toprule
\multirow{2}{*}{Method} & \multicolumn{5}{c}{Accuracy (Normalized Accuracy Gain)}                                                                                                                                      \\ \cmidrule(lr){2-6} 
                        & \multicolumn{1}{c}{gender ($\downarrow$)} & \multicolumn{1}{c}{accent ($\downarrow$)} & \multicolumn{1}{c}{age ($\downarrow$)} & \multicolumn{1}{c}{ID ($\uparrow$)} & digit ($\uparrow$) \\
\midrule
No suppression          & \multicolumn{1}{c}{0.9962 (1.0000)}       & \multicolumn{1}{c}{0.9843 (1.0000)}       & \multicolumn{1}{c}{0.9657 (1.0000)}    & \multicolumn{1}{c}{0.9808 (1.0000)} & 0.9975 (1.0000)    \\
Guessing                & \multicolumn{1}{c}{0.8000 (0.0000)}       & \multicolumn{1}{c}{0.6833 (0.0000)}       & \multicolumn{1}{c}{0.1667 (0.0000)}    & \multicolumn{1}{c}{0.0167 (0.0000)} & 0.1000 (0.0000)    \\
\midrule
ALR                     & \multicolumn{1}{c}{0.7995 (0.0000)}       & \multicolumn{1}{c}{0.6832 (0.0000)}       & \multicolumn{1}{c}{0.1712 (0.0056)}    & \multicolumn{1}{c}{0.6947 (0.7032)} & 0.9970 (0.9994)    \\
GAP                     & \multicolumn{1}{c}{0.8000 (0.0000)}       & \multicolumn{1}{c}{0.6828 (0.0000)}       & \multicolumn{1}{c}{0.1663 (0.0000)}    & \multicolumn{1}{c}{0.6950 (0.7036)} & 0.9597 (0.9579)    \\
MSDA                    & \multicolumn{1}{c}{0.8003 (0.0015)}       & \multicolumn{1}{c}{0.6837 (0.0013)}       & \multicolumn{1}{c}{0.1925 (0.0323)}    & \multicolumn{1}{c}{0.8292 (0.8428)} & 0.9958 (0.9981)    \\
BDQ                     & \multicolumn{1}{c}{0.8000 (0.0000)}       & \multicolumn{1}{c}{0.6835 (0.0007)}       & \multicolumn{1}{c}{0.1677 (0.0013)}    & \multicolumn{1}{c}{0.4060 (0.4038)} & 0.9957 (0.9980)    \\
PPDAR                   & \multicolumn{1}{c}{0.8000 (0.0000)}       & \multicolumn{1}{c}{0.6833 (0.0000)}       & \multicolumn{1}{c}{0.1667 (0.0000)}    & \multicolumn{1}{c}{0.6942 (0.7027)} & 0.9960 (0.9983)    \\ \midrule
MaSS                    & \multicolumn{1}{c}{0.8000 (0.0000)}       & \multicolumn{1}{c}{0.6833 (0.0000)}       & \multicolumn{1}{c}{0.1667 (0.0000)}    & \multicolumn{1}{c}{0.8375 (0.8514)} & 0.9960 (0.9983)    \\
\bottomrule
\end{tabular}
\end{adjustbox}
\end{center}
\end{table}

\begin{table}[!tb]%
\vspace{-0.2in}
\caption{\small Comparison of the classification accuracy and NAG between MaSS and SPAct on AudioMNIST. We suppress gender, accent, age, id, while preserve digit as annotated useful attribute.}
\label{tb:audioMNIST1}
\begin{center}
\begin{adjustbox}{max width=\linewidth}
\begin{tabular}{lccccc}
\toprule
\multirow{2}{*}{Method} & \multicolumn{5}{c}{Accuracy (Normalized Accuracy Gain)}                                                                                                                                        \\ \cmidrule(lr){2-6}
                        & \multicolumn{1}{c}{gender ($\downarrow$)} & \multicolumn{1}{c}{accent ($\downarrow$)} & \multicolumn{1}{c}{age ($\downarrow$)} & \multicolumn{1}{c}{ID ($\downarrow$)} & digit ($\uparrow$) \\ \midrule
No suppression          & \multicolumn{1}{c}{0.9962 (1.0000)}       & \multicolumn{1}{c}{0.9843 (1.0000)}       & \multicolumn{1}{c}{0.9657 (1.0000)}    & \multicolumn{1}{c}{0.9808 (1.0000)}   & 0.9975 (1.0000)    \\
Guessing                & \multicolumn{1}{c}{0.8000 (0.0000)}       & \multicolumn{1}{c}{0.6833 (0.0000)}       & \multicolumn{1}{c}{0.1667 (0.0000)}    & \multicolumn{1}{c}{0.0167 (0.0000)}   & 0.1000 (0.0000)    \\ \midrule
SPAct                   & \multicolumn{1}{c}{0.8087 (0.0442)}       & \multicolumn{1}{c}{0.6833 (0.0001)}       & \multicolumn{1}{c}{0.1753 (0.0108)}    & \multicolumn{1}{c}{0.0707 (0.0560)}   & 0.9948 (0.9970)    \\ \midrule
MaSS                    & \multicolumn{1}{c}{0.8000 (0.0000)}       & \multicolumn{1}{c}{0.6833 (0.0000)}       & \multicolumn{1}{c}{0.1662 (0.0000)}    & \multicolumn{1}{c}{0.0183 (0.0017)}   & 0.9933 (0.9953)    \\
\bottomrule
\end{tabular}
\end{adjustbox}
\end{center}
\end{table}

We also conducted experiments to show the effect of varying the constraint on sensitive attributes suppression ($m$). We take gender, accent, age and ID as sensitive attributes and take digit as annotated useful attribute on the AudioMNIST dataset. We fix $m=0$ for gender, accent and age and $n=2.3$ for digit (its maximal value). Then we vary $m$ for ID from $0$ to $1.46$ (its maximal value). The results are shown in Table \ref{tb:audiovarym}. We can observe that as $m$ increases, the constraint is gradually loosened, which results in the gradually increasing accuracy and NAG for ID.

\begin{table}[!tb]%
\vspace{-0.2in}
\caption{\small Varying the suppression constraint $m$ for ID on AudioMNIST. We suppress gender, accent, age, ID, while preserve digit as if an annotated useful attribute.}
\label{tb:audiovarym}
\begin{center}
\begin{adjustbox}{max width=\linewidth}
\begin{tabular}{lcccccc}
\toprule
\multirow{2}{*}{Method} & \multirow{2}{*}{$m_{ID}$} & \multicolumn{5}{c}{Accuracy (Normalized Accuracy Gain)}                                                                                                                                        \\ \cmidrule(lr){3-7}
                        &                      & \multicolumn{1}{c}{gender ($\downarrow$)} & \multicolumn{1}{c}{accent ($\downarrow$)} & \multicolumn{1}{c}{age ($\downarrow$)} & \multicolumn{1}{c}{ID ($\downarrow$)} & digit ($\uparrow$) \\ \midrule
No suppression          & -                    & 0.9962 (1.0000)                           & 0.9843 (1.0000)                           & 0.9657 (1.0000)                        & 0.9808 (1.0000)                       & 0.9975 (1.0000)    \\
Guessing                & -                    & 0.8000 (0.0000)                           & 0.6833 (0.0000)                           & 0.1667 (0.0000)                        & 0.0167 (0.0000)                       & 0.1000 (0.0000)    \\ \midrule
\multirow{6}{*}{MaSS}   & 0.0                  & 0.8000 (0.0000)                           & 0.6833 (0.0000)                           & 0.1662 (0.0000)                        & 0.0183 (0.0017)                       & 0.9933 (0.9953)    \\
                        & 0.3                  & 0.8000 (0.0000)                           & 0.6833 (0.0000)                           & 0.1665 (0.0000)                        & 0.0598 (0.0447)                       & 0.9938 (0.9959)    \\
                        & 0.6                  & 0.8000 (0.0000)                           & 0.6833 (0.0000)                           & 0.1668 (0.0002)                        & 0.1120 (0.0988)                       & 0.9940 (0.9961)    \\
                        & 0.9                  & 0.8000 (0.0000)                           & 0.6833 (0.0000)                           & 0.1670 (0.0004)                        & 0.1493 (0.1376)                       & 0.9937 (0.9957)    \\
                        & 1.2                  & 0.8000 (0.0000)                           & 0.6833 (0.0000)                           & 0.1667 (0.0000)                        & 0.1963 (0.1863)                       & 0.9928 (0.9948)    \\
                        & 1.46                 & 0.8000 (0.0000)                           & 0.6833 (0.0000)                           & 0.1667 (0.0000)                        & 0.2597 (0.2520)                       & 0.9937 (0.9957)    \\
\bottomrule
\end{tabular}
\end{adjustbox}
\end{center}
\end{table}

Another experiment is to vary constraint $n$ for digit on AudioMNIST, while suppress gender, accent, age, ID with fixed $m=0$. The results are shown in Table \ref{tb:audiovaryn}. We can observe when $n _{digit}$ is large enough, as $n _{digit}$ increases, the constraint posed by annotated attribute preservation module is gradually taken into effect, which gradually turns digit from an unannotated useful attribute (protected by unannotated useful attribute preservation module) to an annotated useful attribute (protected mostly by the annotated useful attribute module), and consequently gradually increases the accuracy and NAG of digit.

\begin{table}[!tb]%
\vspace{-0.2in}
\caption{\small Varying the preservation constraint $n$ for digit on AudioMNIST. We suppress gender, accent, age, ID, while preserve digit as if an
annotated useful attribute.}
\label{tb:audiovaryn}
\begin{center}
\begin{adjustbox}{max width=\linewidth}
\begin{tabular}{lcccccc}
\toprule
\multirow{2}{*}{Method} & \multirow{2}{*}{$n_{digit}$} & \multicolumn{5}{c}{Accuracy (Normalized Accuracy Gain)}                                                                                                                                        \\ \cmidrule(lr){3-7}
                        &                      & \multicolumn{1}{c}{gender ($\downarrow$)} & \multicolumn{1}{c}{accent ($\downarrow$)} & \multicolumn{1}{c}{age ($\downarrow$)} & \multicolumn{1}{c}{ID ($\downarrow$)} & digit ($\uparrow$) \\ \midrule
No suppression          & -                    & 0.9962 (1.0000)                           & 0.9843 (1.0000)                           & 0.9657 (1.0000)                        & 0.9808 (1.0000)                       & 0.9975 (1.0000)    \\
Guessing                & -                    & 0.8000 (0.0000)                           & 0.6833 (0.0000)                           & 0.1667 (0.0000)                        & 0.0167 (0.0000)                       & 0.1000 (0.0000)    \\ \midrule
\multirow{6}{*}{MaSS} & 0.0 & 0.8000 (0.0000) & 0.6833 (0.0000) & 0.1667 (0.0000) & 0.0192 (0.0026) & 0.9685 (0.9677) \\
& 1.8 & 0.8000 (0.0000) & 0.6833 (0.0000) & 0.1657 (0.0000) & 0.0207 (0.0041) & 0.9683 (0.9675) \\
& 1.9 & 0.8000 (0.0000) & 0.6833 (0.0000) & 0.1658 (0.0000) & 0.0178 (0.0012) & 0.9725 (0.9721) \\
& 2.0 & 0.8000 (0.0000) & 0.6833 (0.0000) & 0.1658 (0.0000) & 0.0163 (0.0000) & 0.9733 (0.9731) \\
& 2.1 & 0.8000 (0.0000) & 0.6833 (0.0000) & 0.1642 (0.0000) & 0.0182 (0.0015) & 0.9823 (0.9831) \\
& 2.2 & 0.8000 (0.0000) & 0.6833 (0.0000) & 0.1665 (0.0000) & 0.0202 (0.0036) & 0.9885 (0.9900) \\
& 2.3 & 0.8000 (0.0000) & 0.6833 (0.0000) & 0.1662 (0.0000) & 0.0183 (0.0017) & 0.9933 (0.9953) \\
\bottomrule
\end{tabular}
\end{adjustbox}
\end{center}
\end{table}

\subsection{Evaluation on Facial Images}\label{app:face}

In addition to the experimental results measured in NAG shown in Table \ref{tb:face1}, we also show the experimental results measured in accuracy below in Table \ref{tb:face1acc}.

In the next experiment we demonstrate the performance of MaSS on Adience with different attribute to suppress. We can observe from Table~\ref{tb:face2} that MaSS achieved 0 NAG for suppressed attributes as well as acceptable NAG for preserved unannotated attributes.

\begin{table}[!tb]
    \centering
\caption{\small Comparison of the classification accuracy and NAG between MaSS and baselines on Adience. We suppress gender, while preserve age, ID as if unannotated useful attributes.}
\vspace{0.1in}
\label{tb:face1acc}
\begin{adjustbox}{max width=\linewidth}
\begin{tabular}{llll}
\toprule
\multirow{2}{*}{Method} & \multicolumn{3}{c}{Accuracy (Normalized Accuracy Gain)}                                                                   \\ \cmidrule(lr){2-4} 
                        & \multicolumn{1}{c}{gender ($\downarrow$)} & \multicolumn{1}{c}{age ($\uparrow$)}  & \multicolumn{1}{c}{ID ($\uparrow$)} \\ \midrule
No suppression          & \multicolumn{1}{l}{0.9774 (1.0000)}     & \multicolumn{1}{l}{0.9321   (1.0000)} & 0.9382   (1.0000)                    \\
Guessing                & \multicolumn{1}{l}{0.5240 (0.0000)}       & \multicolumn{1}{l}{0.2892 (0.0000)}   & 0.0284 (0.0000)                      \\ \midrule
ALR                     & \multicolumn{1}{l}{0.5298 (0.0128)}       & \multicolumn{1}{l}{0.2907 (0.0023)}   & 0.0400 (0.0128)                      \\
GAP                     & \multicolumn{1}{l}{0.5240 (0.0000)}       & \multicolumn{1}{l}{0.6047 (0.4907)}   & 0.5393 (0.5616)                      \\
MSDA                    & \multicolumn{1}{l}{0.6652 (0.3114)}       & \multicolumn{1}{l}{0.7989 (0.7928)}   & 0.7982 (0.8461)                      \\
BDQ                     & \multicolumn{1}{l}{0.5252 (0.0026)}       & \multicolumn{1}{l}{0.2892 (0.0000)}   & 0.0352 (0.0075)                      \\
PPDAR                   & \multicolumn{1}{l}{0.5231 (0.0000)}       & \multicolumn{1}{l}{0.2892 (0.0000)}   & 0.0284 (0.0000)                      \\ \midrule
MaSS                    & \multicolumn{1}{l}{0.5240 (0.0000)}       & \multicolumn{1}{l}{0.7661 (0.7418)}   &     0.7255 (0.7662)                 \\\bottomrule
\end{tabular}
\end{adjustbox}
\end{table}

\begin{table}[!tb]
\vspace{-0.1in}
\caption{\small Comparison of the Accuracy and NAG for different configurations of MaSS on Adience. $\checkmark$ denotes that this attribute is suppressed, while all other attributes are preserved as unannotated useful attributes.}
\label{tb:face2}
\begin{center}
\begin{adjustbox}{max width=\linewidth}
\begin{tabular}{l ccc lll}
\toprule
\multirow{2}{*}{Method} & \multicolumn{3}{c}{MaSS Suppressed Attributes}                                                                                                              & \multicolumn{3}{c}{Accuracy (Normalized Accuracy Gain)}                                                                                                      \\ 
\cmidrule(lr){2-4} \cmidrule(lr){5-7} 
                        & \multicolumn{1}{c}{gender}         & \multicolumn{1}{c}{age}          & ID       & \multicolumn{1}{c}{gender} & \multicolumn{1}{c}{age}    &  \multicolumn{1}{c}{ID}  \\ \midrule
No suppression          & \multicolumn{1}{c}{}             & \multicolumn{1}{c}{}               &           & \multicolumn{1}{l}{0.9774 (1.0000)} & \multicolumn{1}{l}{0.9321 (1.0000)}  &    0.9382 (1.0000)                  \\
Guessing                & \multicolumn{1}{c}{$\checkmark$} & \multicolumn{1}{c}{$\checkmark$} & $\checkmark$ & \multicolumn{1}{l}{0.5240 (0.0000)  } & \multicolumn{1}{l}{0.2892 (0.0000)} & 	0.0284 (0.0000)	           \\ \midrule
\multirow{2}{*}{MaSS}   & \multicolumn{1}{c}{$\checkmark$} & \multicolumn{1}{c}{}             &              & \multicolumn{1}{l}{0.5240 (0.0000)}  & \multicolumn{1}{l}{0.7661 (0.7418)} & 	0.7255 (0.7662)	                    \\
                        & \multicolumn{1}{c}{}             & \multicolumn{1}{c}{$\checkmark$} &               & \multicolumn{1}{l}{0.7985 (0.6054)} & \multicolumn{1}{l}{0.2892 (0.0000)} & 0.5005 (0.5189)		  \\\bottomrule       
\end{tabular}
\end{adjustbox}
\end{center}
\end{table}

\begin{figure}[tbp]
    \centering
    \includegraphics[width=1\linewidth]{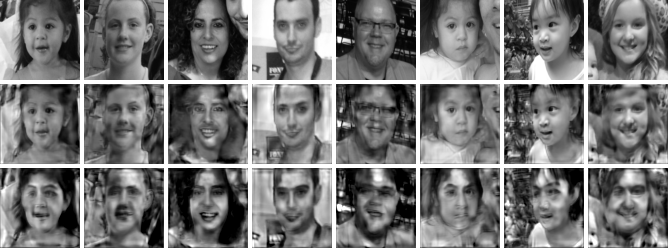}
     \vspace{-0.25in}
    \caption{\small The visualization of the original data and transformed data in Adience dataset. The first row presents the original facial images, while the second and third rows show the transformed images with gender and age suppressed respectively. Other attributes are preserved as unannotated.}
    \label{fig:vis}
     \vspace{-0.1in}
\end{figure}

The visualization results for both original and transformed data in the Adience dataset are depicted in Figure \ref{fig:vis}. Observing the second row, we can see that the gender information has been effectively removed from the images. Similarly, the third row demonstrates the removal of age information from the images, highlighting the efficacy of our approach in suppressing specific attributes.

Although we would not release the labels of sensitive attributes to the public, here we conducted an ablation experiment with the assumption that the attacker can access the ground truth labels of sensitive attributes as an oracle and retrain the discriminator on transformed data. The results are shown in Table 2. We can observe that, using MaSS, the accuracy of the retrained discriminator is higher than adversarial discriminator but is still significantly lower than the discriminator trained using original data. %

\begin{table}[!tb]%
\caption{\small Comparison of the accuracy and NAG between a trained-from-scratch discriminator and adversarial discriminator on the Adience dataset. We suppress gender, while preserve age, ID as if unannotated useful attributes.}
\label{tb:adience1}
\begin{center}
\begin{adjustbox}{max width=\linewidth}
\begin{tabular}{lc}
\toprule
\multirow{2}{*}{Method} & \multicolumn{1}{c}{Accuracy (Normalized Accuracy Gain)} \\ \cmidrule(lr){2-2}
                        & \multicolumn{1}{c}{gender ($\downarrow$)} \\ \midrule
No suppression          & \multicolumn{1}{c}{0.9774 (1.0000)} \\
Guessing                & \multicolumn{1}{c}{0.5240 (0.0000)} \\ \midrule
MaSS (discriminator retrained with oracle) & \multicolumn{1}{c}{0.6029 (0.1740)} \\
MaSS (adversarial discriminator) & \multicolumn{1}{c}{0.5240 (0.0000)} \\
\bottomrule
\end{tabular}
\end{adjustbox}
\end{center}
\end{table}

\subsection{Evaluation on Tabular Marketing Campaign Dataset}\label{app:tabular}
We further evaluate MaSS on the tabular \citet{marketing} dataset and compare the effectiveness of MaSS to MaSS-$\ell_2$. We first convert the categorical attributes into one-hot vectors and normalize the continuous attributes by their ranges.  Note that, the $\ell_2$ reconstruction loss applied  to  one-hot vectors can be interpreted as a 0-1 loss. During training MaSS, we adopt Gumbel-Softmax for the categorical attributes to keep their differentiability and the flexibility to convert them back to the original value. 

Tabular data is slightly different from other data types we experimented in the main paper, the utility and sensitive attributes (columns) are also a part of data $X$ rather than separated attributes.
Thus, we first left out the utility columns as separated attributes ($U$) and train the MaSS over remaining columns to generate a transformed data ($X'$). 
Then, we evaluate the classification accuracy of utility attributes ($U$) using the transformed $X'$; meanwhile, we separate the sensitive column out from the transformed data $X'$ and then using the remaining columns to predict original sensitive columns. The results are shown in Table \ref{tb:tabular}. We can observe that MaSS achieved both higher NAG for response and lower NAG for education compared with MaSS-$\ell_2$, which further validate the generaizability and effectiveness of our framework.

\begin{table}[!tb]
    \centering
\caption{\small Comparison of the classification accuracy and NAG between MaSS and ablation on the Marketing Campaign dataset. We suppress education, while preserve response as annotated useful attribute.}
\vspace{0.1in}
\label{tb:tabular}
\begin{adjustbox}{max width=\linewidth}
\begin{tabular}{lcc}
\toprule
\multirow{2}{*}{Method} & \multicolumn{2}{c}{Accuracy (Normalized Accuracy Gain)}                                                                   \\ \cmidrule(lr){2-3} 
                        & \multicolumn{1}{c}{education ($\downarrow$)} & \multicolumn{1}{c}{response ($\uparrow$)}   \\ \midrule
No suppression          & 0.5223 (1.0000)   & 0.8973 (1.0000)                             \\
Guessing                & 0.4732 (0.0000)      & 0.8504 (0.0000)                            \\ \midrule
MaSS-$\ell_2$           & 0.4933 (0.4094)      & 0.8728 (0.4776)                            \\ \midrule
MaSS                    & 0.4621 (0.0000)     & 0.9084 (1.2367)                \\\bottomrule
\end{tabular}
\end{adjustbox}
\end{table}

\end{document}